\documentclass[11pt,a4paper]{article}
\usepackage[table, dvipsnames]{xcolor}
\usepackage{times,latexsym}
\usepackage{booktabs}
\usepackage{url}
\usepackage[T1]{fontenc}
\usepackage{amsmath}
\usepackage{amssymb}
\usepackage{graphicx}
\usepackage{multicol}
\usepackage{multirow}
\usepackage{todonotes}
\usepackage{fix-cm}
\usepackage{xspace}
\usepackage{enumitem}
\usepackage{mathtools}
\usepackage{graphicx}
\usepackage{multirow}   
\usepackage{algorithm}
\usepackage{amsmath}
\usepackage{algpseudocode}
\usepackage{array}
\usepackage{subfig}
\usepackage{graphicx}
\usepackage{arydshln}
\usepackage{wrapfig}
\usepackage{caption}
\usepackage{soul}
\usepackage{stfloats}
\usepackage{wrapfig}
\usepackage{makecell}
\usepackage{enumitem}

\usepackage[acceptedWithA]{tacl2021v1}
\usepackage{titlesec}
\SetSinglespace{1}
\usepackage{xspace,mfirstuc,tabulary}

\newif\iftaclinstructions
\taclinstructionsfalse 
\iftaclinstructions
\renewcommand{\confidential}{}
\renewcommand{\anonsubtext}{(No author info supplied here, for consistency with
TACL-submission anonymization requirements)}
\newcommand{\instr}
\fi

\titlespacing*{\section}
  {0pt} 
  {10pt} 
  {10pt} 

\titlespacing*{\subsection}
  {0pt} 
  {5pt} 
  {5pt} 

\iftaclpubformat 

\else

\fi

\title{Step-by-Step Unmasking for Parameter-Efficient \\Fine-tuning of Large Language Models}

\author{
   Aradhye Agarwal$^*$\hspace{10pt}
   Suhas Kamasetty Ramesh$^*$\hspace{10pt}
   Ayan Sengupta$^*$\hspace{10pt}
   Tanmoy Chakraborty
   \\
   \\
   Indian Institute of Technology Delhi, India
   \\
   \texttt{Aradhye.Agarwal.cs520@cse.iitd.ac.in, suhaskr@gmail.com}
   \\
   \texttt{ayan.sengupta@ee.iitd.ac.in, tanchak@ee.iitd.ac.in}
 }

\date{}

\makeatletter
\DeclareRobustCommand{\pdot}{\mathbin{\mathpalette\pdot@\relax}}
\newcommand{\pdot@}[2]{%
  \ooalign{%
    $\m@th#1\circ$\cr
    \hidewidth$\m@th#1\cdot$\hidewidth\cr
  }%
}
\makeatother

\def\sparseft{$\texttt{ID}^3$\xspace}
\def\sparseftnospace{$\texttt{ID}^3$}
\def\heuristic{$\texttt{D}^3$\xspace}

\def\gpt3{GPT-3\xspace}

\definecolor{beige}{RGB}{245, 245, 220}

\begin{document}
\maketitle


\begin{abstract}
Fine-tuning large language models (LLMs) on downstream tasks requires substantial computational resources. Selective PEFT, a class of parameter-efficient fine-tuning (PEFT) methodologies, aims to mitigate these computational challenges by selectively fine-tuning only a small fraction of the model parameters. Although parameter-efficient, these techniques often fail to match the performance of fully fine-tuned models, primarily due to inherent biases introduced during parameter selection. Traditional selective PEFT techniques use a fixed set of parameters selected using different importance heuristics, failing to capture parameter importance dynamically and often leading to suboptimal performance. We introduce \sparseft, a novel selective PEFT method that calculates parameter importance continually, and dynamically unmasks parameters by balancing exploration and exploitation in parameter selection. Our empirical study on 16 tasks spanning natural language understanding, mathematical reasoning and summarization demonstrates the effectiveness of our method compared to fixed-masking selective PEFT techniques. We analytically show that \sparseft reduces the number of gradient updates by a factor of two, enhancing computational efficiency. Since \sparseft\ is robust to random initialization of neurons and operates directly on the optimization process, it is highly flexible and can be integrated with existing additive and reparametrization-based PEFT techniques such as adapters and LoRA respectively.\footnote{Code is available at \url{https://github.com/Aradhye2002/selective-peft-toolkit}}
\end{abstract}

\section{Introduction}
\label{sec:intro}

Pre-trained large language models~\cite{devlin2018bert, liu2019roberta, raffel2020exploring, brown2020language, touvron2023llama} have demonstrated remarkable capabilities in understanding and generating natural language. In order to adapt these models to specific downstream tasks, fine-tuning on task-specific datasets is commonly employed to impart specialized domain knowledge. While larger models such as Qwen~\cite{qwen2} and LLaMA~\cite{touvron2023llama} enable promising alternatives like in-context learning (ICL), which allows rapid task adaptation without gradient-based updates, recent research~\cite{liu2022few} suggests that ICL often underperforms compared to fine-tuning in terms of downstream performance and efficiency. As a result, parameter-efficient fine-tuning (PEFT) methods have gained prominence, offering a more practical balance between performance and computational cost when adapting large models to specific tasks.

Parameter-efficient fine-tuning (PEFT) aims to enhance the parameter, memory, and compute efficiency of model fine-tuning by performing low-rank or sparse updates instead of dense updates, as is typical in full fine-tuning (FFT). Additive PEFT methods~\cite{houlsby2019parameter, pfeiffer2020adapterfusion,chen2023hadamard,lei2023conditional} introduce additional trainable parameters to the frozen pre-trained model. In contrast, reparametrization-based PEFT techniques~\cite{hu2021lora, he2022sparseadapter, yang2023bayesian, liu2024dora} utilize low-rank representations of existing model parameters to reduce the number of trainable parameters. Selective methods~\cite{liao2023parameter,sung2021training,zaken2021bitfit,lawton2023neural}, another class of PEFT techniques, use different heuristics to select a subset of parameters within the pre-trained models for fine-tuning. The heuristic function assigns positive real-valued importance to each parameter in the model, while a suitable selection strategy determines which parameters to choose for fine-tuning based on the predicted importance. For instance, Diff Pruning~\cite{guo2020parameter} uses the change in parameter magnitude to assess the parameter importance, whereas Fish mask~\cite{sung2021training} uses a gradient-based Fisher importance heuristic function. Most of these selective PEFT techniques identify and fine-tune only a static set of top-B parameters from the entire parameter pool, where B is a fixed and predefined budget. Incorrect allocation of this budget can detrimentally impact the fine-tuned model's performance due to the suboptimal selection of parameters, either by including non-essential or excluding critical ones.  The parameter selection strategies for these existing selective PEFT techniques can be broadly classified into static (static-S) and repeated (repeat-S). These two strategies represent opposite extremes: static-S is pure exploitation (reusing the same parameters throughout), whereas repeat-S is pure exploration (choosing a fresh set of parameters at each step). A majority of the existing selective PEFT methods use static-S selection, and these exploitation-only methods often fail to select the optimal parameters for a given task. On the other hand, repeat-S-based PEFT methods often overshoot the target budget and perform well only for very small budgets. 

To address these issues, we introduce a novel selection strategy, called increment-S, which balances the exploration and exploitation strategies adopted in repeat-S and static-S, respectively. We analytically show that incremental parameter selection is computationally more efficient and practically beneficial as it provides fine-grained control over the budget, unlike existing methods. Moreover, we experimentally show that despite performing half the number of gradient updates, increment-S performance exceeds existing baselines. We also propose a new \textbf{D}ynamic magnitu\textbf{D}e and gra\textbf{D}ient-based heuristic ({\em aka} \heuristic), which combines the benefits of magnitude and gradient-based parameter importance heuristics. Our proposed method, increment-\heuristic  ({\em aka} \sparseft), can be easily integrated into any neural module and sparsify additive and reparameterized modules of pre-trained models. Existing static-S PEFT techniques do not exhibit this property as they fail to assess parameter importance for randomly initialized untrained parameters. 

\begin{figure}
\centering
    \includegraphics[scale=0.4]{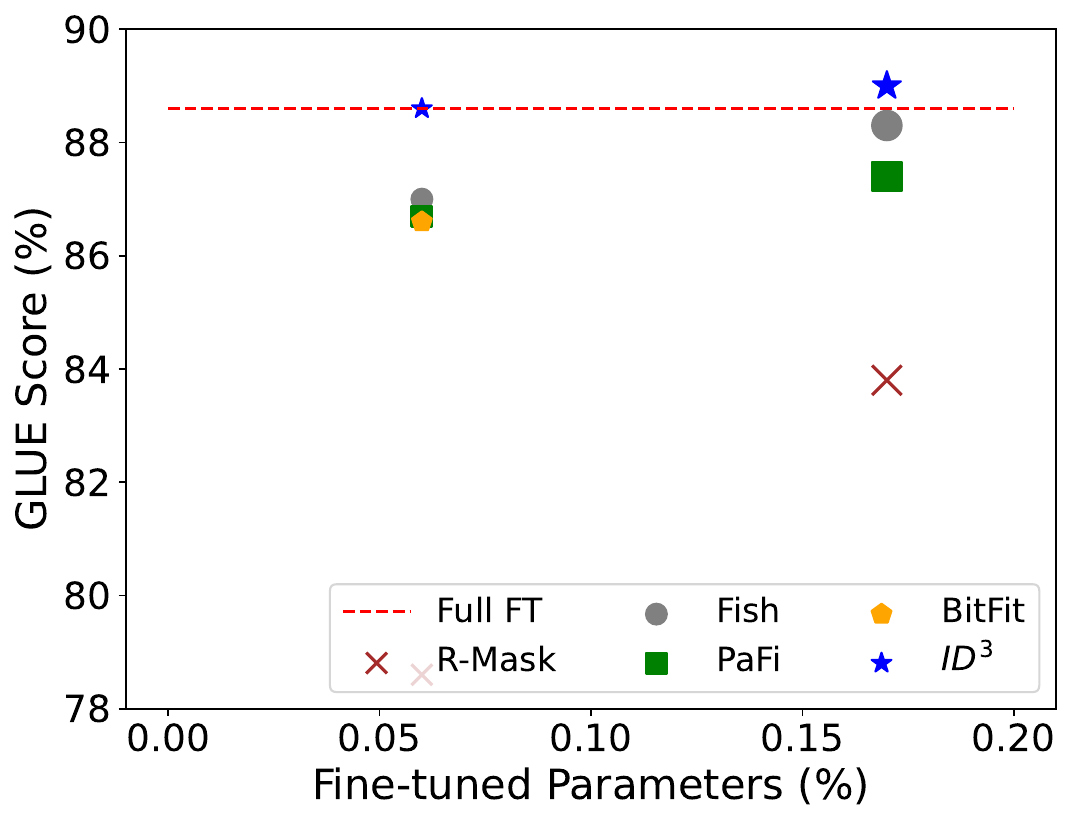}
    \caption{Comparison of different selective PEFT methods -- Full fine-tuning (Full FT), Random masking (R-Mask), Fish~\cite{sung2021training}, BitFit~\cite{zaken2021bitfit}, PaFi~\cite{liao2023parameter} and \sparseft\ on GLUE benchmark. Marker size denotes the number of trainable parameters. Detailed results are reported in Table~\ref{tab:glue_tasks}.}
    \label{fig:glue_results}
\end{figure}

We evaluate the effectiveness of various selective PEFT methods on the GLUE benchmark~\cite{wang2018glue} comprising eight natural language understanding tasks. For a budget of 103K, \sparseft\ outperforms other selective PEFT baselines by a margin of 1.5\% with the pre-trained DeBERTa-v3~\cite{he2021debertav3} backbone. With only 0.17\% of trainable parameters (320K), \sparseft\ beats the fully fine-tuned DeBERTa-v3 model with a margin of 0.45\% on average (c.f.\ Figure~\ref{fig:glue_results}).  We explore \sparseft\ with LoRA~\cite{hu2021lora} fine-tuned LLaMA-7B~\cite{touvron2023llama} and Qwen-2.5~\cite{qwen2} backbone models on six mathematical reasoning tasks, where \sparseft achieves 0.6\% better accuracy than the baselines on zero-shot classification. 

Our major contributions are listed below:
\begin{enumerate}[label=\textbf{(\arabic*)},leftmargin=*,topsep=0pt,itemsep=-1ex,partopsep=1ex,parsep=1ex]
\item We introduce a novel selective strategy, increment-S, for parameter-efficient fine-tuning, which enables incremental parameter selection and dynamic assessment of parameter importance. 
\item We propose a new importance-based heuristic, \heuristic\ that combines the benefits of gradient and magnitude-based parameter importance functions. Together with the increment-S strategy, our proposed selective PEFT method \sparseft demonstrates a strong performance on various natural language understanding and generation tasks, even with highly sparse parameter updates. 
\item Our method produces progressively improved models across increasing budget levels, allowing users to balance budget and performance effectively.
\item We provide an open-source toolkit integrating four selective PEFT techniques, offering comprehensive support for selective methods not available in existing toolkits.
\end{enumerate}

\section{Related Work}
\label{sec:related_works}
This section highlights the representative works in three broad categories of PEFT strategies -- \textit{additive}, \textit{reparameterized} and \textit{selective}.

Additive PEFT methods, such as Adapters~\cite{houlsby2019parameter,pfeiffer2020adapterfusion}, add additional neural components to the pre-trained models. Due to their additive nature, these methodologies usually offer flexibility in multi-task fine-tuning setups, where the same pre-trained model is used with different task-specific adapters. The earliest adapter technique ~\cite{houlsby2019parameter} utilized the additive component in feed-forward networks and attention layers of self-attention~\cite{vaswani2017attention}. Subsequent additive PEFT methods~\cite{he2021towards,li2021prefix,zhu2021counter} differ in terms of placement of these additive components. ~\citet{lei2023conditional} proposed Conditional-Adapter which selectively activates different adapters for different input tokens. ~\citet{chen2023hadamard} came up with a Hadamard Adapter that introduces additional weight and bias parameters and performs element-wise multiplication and addition to the self-attention outputs.

The reparameterization-based PEFT techniques such as LoRA~\cite{hu2021lora} use a low-rank approximation of the parameter update matrix $\Delta W = BA$ to reduce the effective number of trainable parameters. However, LoRA applies a uniform rank across all added parameters, thereby assuming that all parameter matrices are equally important. To address this limitation, AdaLoRA~\cite{zhang2023adaptive} dynamically allocates the parameter budget among the additional weight matrices with singular value decomposition of the $\Delta W$ and importance-aware rank allocation. IncreLoRA~\cite{zhang2023increlora} proposed an incremental parameter allocation method that computes the importance scores of each module and adaptively adds the most important components to the trainable parameters. More recent methods like DyLoRA~\cite{valipour2023dylora}, LoRA+~\cite{hayou2024lora+} and DoRA~\cite{liu2024dora} aim at improving the training efficiency and adaptability of low-rank adaptation on downstream tasks.

Selective parameter-efficient fine-tuning strategies generate a sparse mask $M \in \{0,1\}^{|W|}$ corresponding to each weight matrix $W$ in the pre-trained model. Unlike additive and reparametrization-based techniques, selective methods consider the importance of individual parameters instead of the entire component. In this context, BitFit~\cite{zaken2021bitfit} selectively trains the bias terms within each model parameter. In contrast, Diff Pruning~\cite{guo2020parameter} evaluates the absolute parameter changes across successive training phases, pruning those with the smallest magnitude. Determining the magnitude of parameter change requires significant computational and storage costs, equivalent to full fine-tuning of the model. To alleviate these computational burdens,~\citet{sung2021training, das2023unified} utilized the empirical Fisher importance matrix for selective fine-tuning. To avoid the computation cost of measuring parameter importance, ~\citet{liao2023parameter} proposed PaFi, which assesses the significance based on the absolute magnitude of the parameters and retains only ones with least magnitude. Unlike earlier methods that modify the pre-trained model directly,~\citet{he2022sparseadapter} proposed SparseAdapter, a novel approach that merges with existing adapter-based techniques to sparsify an adapter fine-tuned model, enhancing the efficiency of PEFT. On a similar attempt, ~\citet{zhang2023loraprune} proposed LoRAPrune to combine LoRA with structured pruning to iteratively and progressively reduce model size while maintaining performance.

Our proposed \sparseft\ method distinguishes itself from current selective PEFT methods by progressively selecting the parameters throughout fine-tuning, thereby capturing the change in parameter importance during the training process. Additionally, \sparseft\ can choose model checkpoints with incremental budgets, which is not possible with existing selective PEFT methods. \sparseft\ also leverages both the magnitude and gradient of parameters, which can be efficiently computed using any automatic differentiation tool~\cite{baydin2018automatic}, thereby avoiding extra computational delays.

\section{Methodology}

\begin{figure}[!t]
\captionsetup{font=normalsize}
\centering
    \includegraphics[scale=0.65]{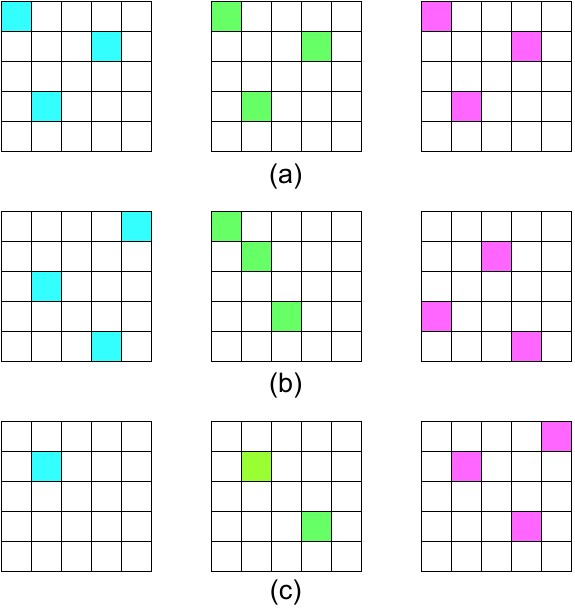}
    \caption{Different parameter selection strategies. Here, $B = 3$ represents the budget while $T = 3$ represents the training steps. \textbf{(a) Static-S} strategy where $B$ number of parameters are chosen initially and used in all future training steps. \textbf{(b) Repeat-S} where $B$ number of fresh parameters are chosen according to the heuristic at each training step. \textbf{(c) Increment-S} where $k = \frac{B}{T}$ parameters are chosen at each training step as per the heuristic.}
    \label{fig:teaser}
\end{figure}

Motivated by the key challenges of the existing selective PEFT methodologies highlighted in Sections~\ref{sec:intro} and \ref{sec:related_works}, we propose \sparseft, an iterative approach for calculating the parameter importance and incrementally selecting the top parameters for each training iteration. We introduce the terms \textit{scalar parameter} and \textit{tensor parameter}, where we refer to individual entries in the weight matrices as scalar parameters and the whole weight matrix itself as the tensor parameter. For instance, a tensor parameter in a BERT~\cite{devlin2018bert} model can be the query matrix of an attention head. The query matrix has $\frac{d^2}{n}$ scalar parameters where $d$ is the hidden dimension, and $n$ is the number of attention heads. We also formulate a selective PEFT method as a heuristic function combined with a selection strategy. We identify three common selection strategies -- (1) Static-S, where the initial set of parameters, selected according to the heuristic, is reused throughout training; (2) repeat-S, where we use the heuristic repeatedly at each training step to find a (potentially) new selected set, and (3) increment-S where we accumulate the selected set over the training iterations, guided by the heuristic. These selection strategies are illustrated in Figure~\ref{fig:teaser}. Existing selective PEFT methods use static-S, \sparseft uses increment-S, while repeat-S is treated as a baseline for comparison.

\subsection{Determining scalar importance}

Evaluating the scalar importance (\textit{i.e.,}\ importance of scalar parameters) of a neural network has always been a pivotal step in model pruning~\cite{molchanov2019importance, cheng2023survey}. For a given neural model, parameterized with $\theta$, we calculate an importance function $f: \mathbb{R}^2 \rightarrow \left[0, \infty\right]$ that measures a real-valued importance for each parameter given its value $\theta^{i}$ and the gradient, $\nabla_{\theta^i}$. Formally, we define the parameter importance function (also referred to as the heuristic function):

\begin{equation}
    \mathcal{H}({\theta^i}) = \frac{|\nabla_{\theta^i}|}{(|\theta^i| + \epsilon)^{exp}},
    \label{eq:heur}
\end{equation}
where $\epsilon \in \left(0, \infty\right)$ and $exp \in \left(-\infty, \infty\right)$ are hyper-parameters to control the smoothing of the function and the effect of parameter magnitude on the final importance respectively. We also note that such a functional form is general enough to represent both the PaFi and Fish metrics by varying the value of $exp$ ($exp = 0$ reduces \heuristic to Fish, while $exp = \infty$ converts it to PaFi). The following theorem also provides the mathematical justification behind the heuristic function.

\noindent \textbf{Definition 1.} Given the output distribution of $y \sim p_{\theta}(\cdot|x)$, where $p_{\theta}(y|x)=f(x,y;\theta)$, for a given input $x$ and a model parameter $\theta$, the Fisher information matrix $\mathcal{I}(\theta)$ is the variance
\begin{multline}
    \mathbb{E}_{x,y} \Bigl[ \Bigl(\frac{\partial}{\partial \theta} \log{f(x,y;\theta)}\Bigr)^2 \Bigl] - \\ \mathbb{E}_{x,y} \Bigl[ \Bigl(\frac{\partial}{\partial \theta} \log{f(x,y;\theta)}\Bigr) \Bigl]^{2}
\end{multline}
Fisher information measures the amount of information the random variable $x$ carries about the unknown model parameter $\theta$ and is widely used to assess the model parameter importance. \\

\noindent \textbf{Theorem 1.} For $\epsilon \geq 1$, $\sqrt{\mathcal{I}(\theta)}$ is the upper bound of $\mathbb{E}_{x,y} \Bigl[ \mathcal{H}(\theta) \Bigr]$. \\
\noindent \textbf{Proof of Theorem 1.} First, we show that $\mathbb{E}_{x,y} \Bigl[ \Bigl(\frac{\partial}{\partial \theta} \log{f(x,y;\theta)}\Bigr) \Bigl] = 0$.
\begin{align*}
    &\mathbb{E}_{x,y} \Bigl[ \Bigl(\frac{\partial}{\partial \theta} \log{f(x,y;\theta)}\Bigr) \Bigl] \\ &= \int_{x}\int_{y} \frac{\frac{\partial}{\partial \theta} {f(x,y;\theta)}}{f(x, y;\theta)} f(x,y;\theta) p(x) \cdot dx \cdot dy\\ &=  \frac{\partial}{\partial \theta}  \int_{x} \int_{y} f(x,y;\theta)p(x) dx \cdot dy = \frac{\partial}{\partial \theta} \cdot 1 = 0
\end{align*}

Therefore, 
\vspace{-7pt}
\begin{align*}
\mathcal{I}(\theta) = \mathbb{E}_{x,y} \Bigl[ \Bigl(\frac{\partial}{\partial \theta} \log{f(x,y;\theta)}\Bigr)^2 \Bigl]
\end{align*}

\vspace{-20pt}
\begin{align*}
\mathbb{E}_{x,y} \Bigl[ \mathcal{H}(\theta) \Bigr]  = \frac{ \mathbb{E}_{x,y} \Bigl[ \Bigl| \frac{\partial}{\partial \theta} \log{f(x,y;\theta)} \Bigr|\Bigr]}{(|\theta| + \epsilon)^{exp}}
\end{align*}

Using Jensen's inequality, we get, 
\vspace{-15pt}

\begin{align*}
    \mathcal{I}(\theta) &= \mathbb{E}_{x,y} \Bigl[ \Bigl| \frac{\partial}{\partial \theta} \log{f(x,y;\theta)} \Bigr|^{2} \Bigr] \\ &\geq \Bigl(  \mathbb{E}_{x,y} \Bigl[  \Bigl| \frac{\partial}{\partial \theta} \log{f(x,y;\theta)} \Bigr| \Bigr] \Bigr)^{2} \\
    &= \Bigl( \mathbb{E}_{x,y} \Bigl[ \mathcal{H}(\theta) \Bigr] \Bigr)^{2} \cdot (|\theta| + \epsilon)^{2 \cdot exp}
\end{align*}

Hence, for $\epsilon \geq 1$ and $exp \geq 0$,  $\mathcal{I}(\theta) \geq \Bigl( \mathbb{E}_{x,y} \Bigl[ \mathcal{H}(\theta) \Bigr] \Bigr)^{2} $. Therefore, Theorem 1 justifies that maximizing $\mathcal{H}(\theta^{i})$ indirectly maximizes Fisher importance.

\begin{algorithm}\small
\caption{Incremental parameter updates}\label{alg:cap}
\begin{algorithmic}
\Require  Unmasking scheduler $\{u_t\}_{t=1}^T$, number of training steps $T$, trainable model $\theta_{(0)}$, training dataset $(X,Y)$, learning rate $\eta$
\State $t \gets 0$
\State $\Lambda_{0} \gets \emptyset$
\While{$t < T$}
\State $(x,y) \sim (X,Y)$ minibatch
\State Compute predicted output $\hat{y} = p_{\theta_{(t)}}(\cdot|x)$
\State Compute loss $l = \mathcal{L}\left(y, \hat{y}\right)$
\State Compute gradient $\nabla_{\theta_{(t)}} = \nabla_{\theta_{(t)}}l$
\State Compute parameter importance $\mathcal{H}$ for parameters in $\theta_{(t)} \setminus \Lambda_t$ using Equation~\ref{eq:heur}
\State Find scalar parameters $\lambda_{t}$ using Equation~\ref{eq:incremental_selection}
\State $\Lambda_{t+1} \gets \Lambda_{t} \cup \lambda_t$
\State Update parameter gradients $\tilde{\nabla}_{\theta_{(t)}}$ using Equation~\ref{eq:gradient_update}
\State Perform parameter update $\theta_{(t+1)} \gets \theta_{(t)} + \eta \tilde{\nabla}_{\theta_{(t)}}$
\State $t \gets t+1$
\EndWhile
\end{algorithmic}
\end{algorithm}

\subsection{Incremental parameter updates}

Suppose we want to fine-tune a pre-trained model parameterized by $\theta_{(0)}$ (0 denotes the fine-tuning timestep), with $|\theta_{(0)}| = N$ on a task for maximum $T$ number of steps. Suppose we fix the budget of fine-tuning as $B$, \textit{i.e.,} we only fine-tune a maximum of $B$ number of scalar parameters in the entire model training. The factor $\frac{N-B}{N}$ is called \textit{sparsity} of the model. We choose a suitable unmasking scheduler $\{u_t\}^{T}_{t=1}$ that estimates the number of parameters to be updated in each iteration $t$. By default, we use a uniform scheduler where $u_t = \frac{B}{T}$. At the beginning of model fine-tuning, the unmasked parameters $\Lambda_{t} = \emptyset$. At each training iteration $t$, we measure the importance for each parameter in the set $\theta_{(t-1)} \setminus \Lambda_{t-1}$ using Equation~\ref{eq:heur} and determine the incremental unmasked parameters $\Lambda_t$ such that

\begin{equation}
    \max_{\lambda_t} \min_{\theta^i \in \lambda_t} \{\mathcal{H}(\theta^i) \} \text{ s.t. } |\lambda_t| = u_t 
    \label{eq:incremental_selection}
\end{equation}
Finally, the set of unmasked parameters is updated as $\Lambda_{t} = \Lambda_{t-1} \cup \lambda_{t}$. During the forward pass, we compute the task-specific loss $\mathcal{L}\Bigl(y, p_{\theta_{(t)}}(\cdot|x)\Bigr)$, while during the backward pass, the gradients $\nabla_{\theta_{(t)}}$ are set to zeros for parameters not in the unmask set $\Lambda_t$, obtaining $\tilde{\nabla}_{\theta_{t}}$. Formally,

\begin{equation}\small
  \tilde{\nabla}_{\theta^{i}_{t}} = \left \{
  \begin{aligned}
    &\nabla_{\theta^{i}_{t}}, && \text{if}\ \theta^{i}_{t} \in \Lambda_t \\
    &0 && \text{otherwise}\\
  \end{aligned} \right.
  \label{eq:gradient_update}
\end{equation} 
Finally, the parameters are updated using the filtered gradients $\tilde{\nabla}_{\theta_{(t)}}$. Algorithm~\ref{alg:cap} formalizes the \sparseft incremental parameter update procedure. With the incremental parameter selection and updates, the total number of parameter updates can be calculated as$$U_{dynamic} = \sum_{t=0}^{T-1}\sum_{i=0}^t u_i.$$ For the uniform unmasking scheduler, 
$$U_{dynamic} = \sum_{t=0}^{T-1}\sum_{i=0}^t \frac{B}{T}  = \frac{T+1}{2}B.$$ For static-masking-based PEFT techniques, the total number of parameter updates is $$U_{static} = \sum_{t=0}^TB = T\cdot B$$ 
Hence, 
$$U_{dynamic} = \frac{U_{static}}{2} \hspace{10pt}\text{(when $T \gg 1$)}.$$ Therefore, incremental selection with a uniform schedule can reduce the effective number of gradient updates by a factor of 2. 

\begin{figure}[t]
\centering
    \includegraphics[scale=0.65]{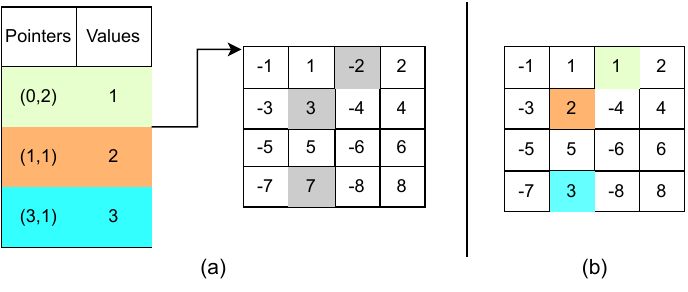}
    \caption{\textbf{(a, right):} Tensor parameter in the pre-trained model. \textbf{(a, left):} Table storing pointers and corresponding values of the scalar parameters updated during fine-tuning. \textbf{(b):} Final tensor parameter in the fine-tuned model, where old scalar values are replaced with updated ones.}
    \label{fig:main_fig}
\end{figure}

\subsection{Efficient processing of sparse masks}
Storing and loading the sparse masks requires efficient handling of the masked scalar parameters. For storing the sparse weights, we store only the weights of the unmasked scalar parameters and their corresponding pointers. Since the maximum dimension of any tensor does not typically exceed 2, we need to store at most two indices for any given scalar parameter, which can be stored using a 32-bit unsigned integer. Each updated model parameter, on the other hand, can be stored using 64-bit double floating point numbers. Therefore, we can reduce the space complexity required to just $\mathcal{O}(2\times32\times B+64\times B) = \mathcal{O}(B)$. While loading, we can use these pointers (stored in the form of tensors) to index into the tensor parameters and replace the pre-trained parameters with the stored ones that were learned during selective fine-tuning.  Figure~\ref{fig:main_fig} summarizes the process of handling sparse masks.

\section{Experimental Setup}

\subsection{Datasets and tasks}

To evaluate the effectiveness of our proposed method, we conduct exhaustive experiments across three distinct tasks: text classification, token classification, and text generation.

For text classification, we use eight tasks from the GLUE benchmark~\cite{wang2018glue}: CoLA, MRPC, RTE, STS-B, SST-2, MNLI-m/mm, QNLI, and QQP. In line with previous studies~\cite{liao2023parameter, sung2021training, zaken2021bitfit}, we exclude the WNLI task due to its poor performance with pre-trained language models. On token classification, we experiment with the named entity recognition (NER) task using the CoNLL-2003 dataset~\cite{tjong-kim-sang-de-meulder-2003-introduction}. For these nine tasks, we fine-tune the model using the training splits and evaluate its performance on the validation splits.

For text generation we consider the CNN/Daily Mail summarization~\cite{hermann2015teaching, nallapati2016abstractive} task and six arithmetic reasoning tasks: GSM8K, SVAMP, MultiArith, AddSub, AQuA, and SingleEq. For the summarization task we train and evaluate on the training and dev split of the original dataset. For the arithmetic reasoning tasks we fine-tune the models on the Math10K dataset, as curated by~\citet{hu2023llm}, and evaluate them on the test splits of the datasets above. We use 10\% of the training data for validation. Detailed descriptions of these datasets and tasks are provided in Section~\ref{app:datasets} and Table~\ref{tab:datasets_splits} of Appendix.

\subsection{Models} For NLU and NER tasks, we use the pre-trained encoder-only DeBERTa-v3-base ~\cite{he2021debertav3} 
 and RoBERTa-base~\cite{liu2019roberta} models as the backbone, while for summarization, we use the pre-trained T5-small~\cite{raffel2020exploring} model. For math reasoning tasks, we use pre-trained LLaMA-7B~\cite{touvron2023llama}, Qwen-2.5~\cite{qwen2} and MobileLLaMA\mbox{-}2.7B~\cite{chu2023mobilevlm} models. All the pre-trained model weights are obtained from Huggingface~\cite{wolf-etal-2020-transformers}. 

\subsection{Toolkit implementation} 
A significant contribution of our work is the implementation of the \textit{selective-peft-toolkit}\footnote{The toolkit will be open-sourced upon acceptance.}. We use PyTorch ~\cite{NEURIPS2019_9015} and the huggingface transformers library ~\cite{wolf-etal-2020-transformers} for implementing the toolkit. We implement the following selective PEFT baselines in our toolkit: (1) BitFit~\cite{zaken2021bitfit} which involves fine-tuning only the bias terms in a pre-trained model;
(2) PaFi~\cite{liao2023parameter}  which selects the pre-trained parameters with the smallest magnitude and trains only these parameters during fine-tuning;
(3) \sparseft.

The toolkit allows integration of these selective PEFT methods into the original pre-trained models as well as into any additional neural modules such as Adapters~\cite{houlsby2019parameter, pfeiffer2020adapterfusion} and LoRA~\cite{hu2021lora}. We also provide methods for storing and loading the sparse weights in a memory efficient manner, enabling end-to-end training and evaluation workflows.

Additional details and hyperparameters for reproducing the results are provided in Section~\ref{app:hyperparameters} and Table~\ref{tab:hyperparameters_all} of the Appendix. All experiments are conducted on Nvidia A100 and A6000 GPUs.

\section{Results}
This section presents the results of our exhaustive experiments on text classification, token classification and text generation.
\label{sec:results}
\subsection{Text classification}

\begin{table*}[!ht]
    \centering
    \resizebox{2.08\columnwidth}{!}{
    \begin{tabular}{llcccccccccc}
        \toprule
        Budget & Method & MNLI-m & MNLI-mm & QQP & QNLI & SST-2 & STS-B & CoLA & MRPC & RTE & Avg \\
        \midrule
        184M & Full-FT & 90.21 & 90.32 & 91.98 & 94.1 & 96.16 & 90.89 & 70.65 & 89.71 & 83.21 & 88.58 \\
        \hdashline\noalign{\vskip 0.5ex}
        \multirow{5}{*}{103K} 
         & R-Mask & 82.74 & 83.63 & 86.11 & 88.04 & 92.77 & 80.66 & 60.24 & 76.04 & 57.22 & 78.61 \\
         & Fish & 87.91 & 88.11 & 87.35 & 92.09 & 95.10 & 91.30 & 68.12 & 90.01 & 83.31 & 87.03 \\
         & PaFi & 87.80 & 87.99 & 88.97 & 93.20 & 95.53 & 89.12 & 67.67 & 89.34 & 80.69 & 86.70 \\
         & BitFit & 88.10 & 88.03 & 88.61 & 92.83 & 95.13 & 89.11 & 68.75 & 89.10 & 79.88 & 86.61 \\
         & \sparseft & \textbf{89.33} & \textbf{89.59} & \textbf{89.84} & \textbf{93.62} & \textbf{95.56} & \textbf{91.97} & \textbf{70.49} & \textbf{90.81} & \textbf{85.83} & \textbf{88.56} \\
        \hdashline\noalign{\vskip 0.5ex}
        \multirow{4}{*}{320K} 
         & R-Mask & 87.32 & 87.54 & 88.47 & 91.35 & 94.67 & 85.61 & 64.84 & 81.68 & 72.38 & 83.76 \\
         & Fish & 88.94 & 89.66 & 88.73 & 93.93 & 95.53 & 91.92 & 69.25 & 90.57 & 86.64 & 88.35 \\
         & PaFi & 89.15 & 89.27 & 89.97 & 93.71 & 95.84 & 89.84 & 68.39 & 90.20 & 80.60 & 87.44 \\
         & \sparseft & \textbf{89.58} & \textbf{89.73} & \textbf{90.31} & \textbf{94.03} & \textbf{95.90} & \textbf{91.97} & \textbf{71.46} & \textbf{91.12} & \textbf{87.19} & \textbf{89.03} \\
         \rowcolor{beige}
        \multicolumn{2}{c}{Wilcoxon statistic} & \bf 465.0  & \bf 474.0  & \bf 459.5 & \bf 400.5  & 234.5  & \bf 402.0  & \bf 525.5  & \bf 389.5  & \bf 359.0  & \bf 493.0  \\
        \rowcolor{beige}
        \multicolumn{2}{c}{(p-value)} & \bf (1e-5) & \bf (1e-5) & \bf (5e-5) & \bf (1e-3) & (0.71) & \bf  (4e-3) & \bf (1e-9) & \bf (9e-5) & \bf (1e-2) & \bf (8e-7) \\
        \bottomrule
    \end{tabular}}
    \caption{
    Mean performance of selective PEFT methods on GLUE tasks with DeBERTa-v3. BitFit is evaluated only at the 103K budget, corresponding to DeBERTa-v3's 103K bias parameters. R-mask denotes a random static mask baseline. The best-performing method within each budget group is shown in \textbf{bold}. Standard deviations are provided in Table~\ref{tab:std_glue} of Appendix~\ref{app:std_deviations}. We calculate the Wilcoxon statistic (and the associated p-value) for each GLUE task to assess the statistical significance of the improvement shown by \sparseft over the baselines. \textbf{Bold} indicates the tasks where p-value < 0.05 . 
    \vspace{7pt}
    }
    \label{tab:glue_tasks}
\end{table*}

\begin{table*}[!htb]
\captionsetup{font=normalsize}
\centering
\resizebox{2.08\columnwidth}{!}{%
\begin{tabular}{llcccccccccc}
\toprule
Budget & Method             & MNLI-m  & MNLI-mm & QQP   & QNLI  & SST-2   & STS-B  & CoLA    & MRPC    & RTE    & Avg    \\
\midrule
1.33M & LoRA (r=8)          & 90.47   & 90.46   & 91.95 & 93.76 & 95.57   & 91.86  & 69.73   & 89.71   & 85.32  & 88.76  \\
\hdashline\noalign{\vskip 0.5ex}
320K & PaFi + LoRA (r=8)    & 89.95 & 89.89 & 91.20 & 94.09 & \textbf{95.99} & 90.90 & \textbf{70.22} & 90.01 & 83.87 & 88.46 \\
320K & \sparseft + LoRA (r=8) & \textbf{90.11} & \textbf{90.06} & \textbf{91.48} & \textbf{94.15} & 95.70 & \textbf{91.58} & 68.83 & \textbf{90.50} & \textbf{86.46} & \textbf{88.76} \\
\rowcolor{beige}
\multicolumn{2}{c}{Wilcoxon statistic} & 83.0 & 95.0 & 82.0 & 61.0 & 0.0 & \textbf{134.0} & 1.0 & \textbf{77.0} & \textbf{131.5} & \textbf{103.0} \\
\rowcolor{beige}
\multicolumn{2}{c}{(p-value)}          & (0.09) & (0.09) & (0.25) & (0.65) & (0.99) & \textbf{(5e-5)} & (0.99) & \textbf{(0.01)} & \textbf{(1e-4)} & \textbf{(0.04)} \\
\bottomrule
\end{tabular}%
}
\caption{Performance of PaFi and \sparseft with LoRA+pretrained DeBERTa-v3 on GLUE tasks. Standard deviations are reported in Table~\ref{tab:std_glue_lora} (Appendix~\ref{app:std_deviations}). The average improvement of \sparseft over PaFi is statistically significant, but results remain inconclusive for six of nine tasks (p-value $\ge $ 0.05).}
\label{tab:glue_lora}
\end{table*}

We report the results on GLUE tasks in Table~\ref{tab:glue_tasks}. \sparseft achieves an average score of 89.03\% with a budget of 320K, surpassing the best-performing baseline (Fish) by over 0.6\%. Interestingly we observe that \sparseft outperforms even the FFT baseline (88.58\%). A similar comparison holds at the smaller budget level of 103K, with \sparseft outperforming other selective baselines by more than 1\%.  We perform paired Wilcoxon tests\footnote{Additional details regarding the significance testing methodologies are presented in Appendix~\ref{appx:stat_test}.} between the results obtained by \sparseft and the best baselines (for each task) across all the budgets to compute the Wilcoxon statistic. At an overall level, we obtain a Wilcoxon statistic of 103.0 with a p-value of 0.04, indicating the statistical significance of the competitive performance of \sparseft. \sparseft outperforms existing baselines with statistical significance on 8 of 9 GLUE tasks.

\begin{table}[!t]
\captionsetup{font=normalsize}
\centering
\resizebox{\columnwidth}{!}{%
\begin{tabular}{llccccc}
\toprule
Budget & Method            & STS-B & CoLA & MRPC & RTE & Average       \\ \midrule
8M & Pfeiffer & 90.78 & 59.05 & 89.21 & 76.53 & 78.89 \\ \hdashline\noalign{\vskip 0.5ex}
320K & \makecell{SparseAdapter \\ + Pfeiffer} & \bf 90.88 & 58.95 & 89.41 & 77.03 & 79.07 \\
320K & \sparseft\ + Pfeiffer & 90.71 & \bf 59.84 & \bf 89.95 & \bf 79.42 & \bf 79.98\\
\bottomrule
\end{tabular}
}
\caption{Performance of \sparseft compared with SparseAdapter~\cite{he2022sparseadapter} on Pfeiffer adapter~\cite{pfeiffer2020adapterfusion} applied to pretrained RoBERTa~\cite{liu2019roberta}. A Wilcoxon statistic of 9.0 highlights that \sparseft outperforms SparseAdapter, however, a p-value of 0.12 indicates that the results cannot be concluded statistically significant under a significance level of 0.05.}
\label{tab:reused_mask_results}
\end{table}

We further evaluate the effectiveness of \sparseft with other adapters integrated with pre-trained language models. Table~\ref{tab:glue_lora} reports the performance of the DeBERTa-v3 model with rank 8 (indicated by r=8) LoRA adapter, with and without \sparseft. With a budget of 320K (sparsity 76\%), \sparseft matches full LoRA fine-tuning with an average of 88.76\%. Interestingly, LoRA sparsified with both \sparseft\ and PaFi beats the dense LoRA model on four of nine GLUE tasks, indicating the importance of sparsification of adapters for more efficient and effective fine-tuning. An empirical study with adapters~\cite{pfeiffer2020adapterfusion} narrates a similar phenomenon as shown in Table~\ref{tab:reused_mask_results}. With a budget of only 320K (sparsity 96\%), \sparseft can improve the performance of an adapter-integrated RoBERTa-base by a margin of 1.09\%. SparseAdapter, another popular sparsification technique for adapters, falls short by 0.91\% compared to \sparseft.

\subsection{Token classification}

\begin{table}[!t]
    \captionsetup{font=normalsize}
    \centering
    \resizebox{1\columnwidth}{!}{
    \begin{tabular}{lcccccc}
        \toprule
        Budget & Full-FT & Fish & PaFi & BitFit & \sparseft & R-Mask \\ \midrule
        103K & - & 95.26 & 94.40 & 93.85 & \textbf{95.55} & 70.15 \\ 
        320K & - & 95.93 & 95.42 & - & \textbf{96.04} & 89.93 \\ 
        184M & 96.62 & \multicolumn{5}{c}{-} \\ 
        \bottomrule
    \end{tabular}}
    \caption{Mean performance of selective methods with DeBERTa-v3 on NER at different budgets. As the DeBERTa model has 103K bias terms, BitFit is only run with the 103K budget. Corresponding standard deviations are reported in Table~\ref{tab:std_ner} of Appendix~\ref{app:std_deviations}. A Wilcoxon statistic of 385.0 with p-value 0.01 indicates the statistical significance of improvement shown by \sparseft over the baselines}.
    \label{tab:ner_2}
    
\end{table}
\begin{table}[!t]
\captionsetup{font=normalsize}
\centering
\resizebox{0.99\columnwidth}{!}{
\begin{tabular}{lcccc}
\toprule
Budget & Method & Rouge-1 & Rouge-2 & Rouge-L \\
\midrule
60M & FFT & 41.29 & 18.90 & 29.19 \\
\hdashline\noalign{\vskip 0.5ex}
\multirow{2}{*}{100K} & PaFi       & 40.15 & 18.03 & 28.49 \\
                      & \sparseft  & \textbf{40.43} & \textbf{18.44} & \textbf{28.76} \\
\hdashline\noalign{\vskip 0.5ex}
\multirow{2}{*}{320K} & PaFi       & 40.75 & 18.57 & 28.83 \\
                      & \sparseft  & \textbf{40.91} & \textbf{18.73} & \textbf{28.98} \\
\hdashline\noalign{\vskip 0.5ex}
\multirow{2}{*}{1M}   & PaFi       & 41.16 & 18.79 & 29.09 \\
                      & \sparseft  & \textbf{41.17} & \textbf{18.85} & \textbf{29.17} \\
\rowcolor{beige}
\multicolumn{2}{c}{Wilcoxon statistic} & 6.0 & 6.0 & 6.0 \\
\rowcolor{beige}
\multicolumn{2}{c}{(p-value)}          & (0.13) & (0.13) & (0.13) \\
\bottomrule
\end{tabular}}
\caption{Performance of \sparseft and PaFi with T5-small on summarization.}
\label{tab:summ_mean}
\end{table}

The CoNLL benchmark results in Table~\ref{tab:ner_2} highlight \sparseft as a top-performing PEFT method, achieving an F1 score of 95.55\% with only 103K parameters surpassing Fish (95.26\%) and PaFi (95.40\%). With a larger 320K parameter budget, \sparseft improves to 96.04\%, approaching FFT's baseline of 96.62\% (184M parameters). This demonstrates \sparseft's efficiency and robustness as a highly effective alternative to full fine-tuning.

\subsection{Text generation}
We evaluate \sparseft along with the other selective baselines on two text generation tasks which include abstractive summarization and mathematical reasoning.
\subsubsection{Summarization}

\begin{table*}[t]
\captionsetup{font=normalsize}
\centering
\resizebox{2.08\columnwidth}{!}{
\begin{tabular}{lllccccccc}
\toprule
Model & Budget & Method & AddSub & MultiArith & SingleEq & GSM8K & AQuA & SVAMP & Avg. \\
\midrule
\multirow{4}{*}{LLaMA-7B} & 56M & LoRA (r=32)              & 81.3 & 95.5 & 81.7 & 34.1 & 17.7 & 46.7 & 59.5 \\
\cdashline{2-10}\noalign{\vskip 0.5ex}
& 3.5M & LoRA (r=2)               & 78.2 & \bf 96.7 & 76.6 & \bf 35.3 & \bf 16.9 & 44.9 & 58.1 \\
& 3.5M & PaFi + LoRA (r=32) &           78.7 & 92.3 & 76.8 & 33.9 & \bf 16.9 & 43.2 & 57.0 \\
& 3.5M & \sparseft + LoRA (r=32) &           \bf 80.7 & 95.8 & \bf 79.3 & 34.3 & 15.7 & \bf 45.7 & \bf 58.6 \\
\midrule
\multirow{4}[0]{*}{Qwen-7B} & 54M   & LoRA (r=32) & 94.4  & 98.2  & 97.6  & 76.9  & 34.6  & 85.8  & 81.3 \\
\cdashline{2-10}\noalign{\vskip 0.5ex}
& 3.4M  & LoRA (r=2) & \bf 93.9  & 98.3  & 96.4  & 76.4  & 31.9  & 86.8  & 80.6 \\
& 3.4M  & PaFi + LoRA (r=32) & 91.1  & \bf 99.0    & \bf 97.0    & \bf 78.5  & \bf 37.8  & 85.8  & \bf 81.5 \\
& 3.4M  & \sparseft + LoRA (r=32) & 93.6  & 98.5  & 95.1  & 77.9  & 37.0    & \bf 87.1  & \bf 81.5 \\
\midrule
\multirow{4}[0]{*}{Qwen-3B} & 40M   & LoRA (r=32) & 92.1  & 98.5  & 95.9  & 71.9  & 34.2  & 81.5  & 79.0 \\
\cdashline{2-10}\noalign{\vskip 0.5ex}
& 2.5M  & LoRA (r=2) & \bf 92.9  & 97.5  & 94.9  & 70.8  & 34.6  & \bf 85.1  & 79.3 \\
& 2.5M  & PaFi + LoRA (r=32) & 90.9  & 97.8  & \bf 96.2  & 70.6  & 36.2  & 83.9  & 79.3 \\
& 2.5M  & \sparseft + LoRA (r=32) & 92.6  & \bf 98.2  & 95.9  & \bf 71.5  & \bf 37.4  & 83.9  & \bf 79.9 \\
\midrule
\multirow{4}[0]{*}{Qwen-1.5B} & 25M   & LoRA (r=32) & 90.4  & 98.2  & 96.6  & 65.8  & 36.6  & 75.3  & 77.2 \\
\cdashline{2-10}\noalign{\vskip 0.5ex}
& 1.5M  & LoRA (r=2) & \bf 91.9  & \bf 98.2  & 95.5  & 62.8  & 31.1  & 80.9  & 76.7 \\
& 1.5M  & PaFi + LoRA (r=32) & 89.4  & 96.7  & \bf 95.9  & \bf 64.5  & 32.3  & 78.4  & 76.2 \\
& 1.5M  & \sparseft + LoRA (r=32) & 91.6  & 97.8  & 93.7  & 62.6  & \bf 34.6  & \bf 81.0    & \bf 76.9 \\
\rowcolor{beige}
\multicolumn{3}{c}{Wilcoxon statistic} & 4.0 & 4.0 & 4.0 & 4.0 & 6.0 & 6.0 & 10.0 \\
\rowcolor{beige}
\multicolumn{3}{c}{(p-value)}          & (0.69) & (0.69) & (0.69) & (0.69) & (0.44) & (0.44) & (0.06) \\

\bottomrule
    \end{tabular}}
\caption{Results on mathematical reasoning obtained from LLaMA and Qwen with LoRA fine-tuning. We report the Wilcoxon statistic alongside the associated p-value for highlighting the statistical significance of the results.} 
\vspace{10pt}
\label{table:generative_tasks}
\end{table*}

\begin{table*}[h]
\captionsetup{font=normalsize}
\centering
\resizebox{2.08\columnwidth}{!}{
\begin{tabular}{lllccccccc}
\toprule
Model & Budget & Method & AddSub & MultiArith & SingleEq & GSM8K & AQuA & SVAMP & Avg. \\
\midrule
\multirow{5}[0]{*}{MobileLLaMA-2.7B} & 2.7B  & FFT   & 79.7  & 95.8  & 82.2  & 33.3  & 18.1  & 31.8  & 56.8 \\
\cdashline{2-10}\noalign{\vskip 0.5ex}
& 2.7M  & PaFi  & 46.1  & 66.5  & 46.6  & 11.1  & \bf 18.7  & 23.1  & 35.4 \\
& 2.7M  & \sparseft   & \bf 47.1  & \bf 67.8  & \bf 48.4  & \bf 11.9  & 17.3  & \bf 25.0    & \bf 36.3 \\
\cdashline{2-10}\noalign{\vskip 0.5ex}    
& 1.3M  & PaFi  & 30.1  & 36.2  & 30.7  & 8.1   & \bf 21.2  & 17.5  & 24.0 \\
& 1.3M  & \sparseft   & \bf 35.2  & \bf 57.8  & \bf 41.1  & \bf 8.6   & 15.7  & \bf 22.0    & \bf 30.1 \\
\bottomrule
\end{tabular}}
\caption{Results of MobileLLaMA-2.7B with full fine-tuning on mathematical reasoning tasks. A Wilcoxon statistic of 88.0 with a p-value of 0.01 indicates the statistical significance of \sparseft's improvement over PaFi.}
\label{table:generative_tasks2}
\end{table*}

The results of T5-small on the CNN/Daily Mail summarization task in Table~\ref{tab:summ_mean} show that fine-tuning all 60M parameters (FFT) achieves the highest performance with Rouge-1 of 41.29, Rouge-2 of 18.90, and Rouge-L of 29.19. Among the selective methods, \sparseft consistently outperforms PaFi across all parameter budgets. At 100K parameters, \sparseft achieves Rouge scores of 40.43/18.44/28.76, improving over PaFi by 0.28/0.41/0.27 points. At 320K, \sparseft improves to 40.91/18.73/28.98, surpassing PaFi by 0.16/0.16/0.15. At 1M, \sparseft scores 41.17/18.85/29.17, slightly outperforming PaFi. While FFT remains superior, \sparseft demonstrates its efficiency and robustness as an effective alternative under constrained parameter budgets. 

\subsubsection{Mathematical reasoning}

Table~\ref{table:generative_tasks} presents the results of various mathematical reasoning tasks. LLaMA-7B fine-tuned with LoRA (r=32) achieves a strong baseline average score of 59.5\%. Notably, even when the parameter budget is reduced to 3.5M, LoRA (r=2) maintains robust performance with an average of 58.1\%, excelling in MultiArith (96.7\%) but showing minor drops on other tasks compared to the 56M setting. Applying \sparseft to LoRA (r=32) yields a slightly higher average score of 58.6\%, outperforming LoRA (r=2) with the same parameter budget. This setup delivers strong results on AddSub (80.7\%) and SingleEq (79.3\%), suggesting that sparsifying higher-rank LoRA modules enhances performance. PaFi combined with LoRA achieves an average score of 57.0\%, with its best result in MultiArith (92.3\%), though it generally trails behind both full-rank LoRA and \sparseft in other tasks. On Qwen-7B, both PaFi + LoRA and \sparseft + LoRA reach an average score of 81.5, marginally surpassing LoRA (r=32) at 81.3. Similar trends hold for Qwen-3B and Qwen-1.5B, where \sparseft + LoRA consistently matches or exceeds the performance of PaFi + LoRA while maintaining parameter efficiency. Specifically, \sparseft leads in reasoning-heavy tasks like GSM8K (71.5 vs.\ 70.6 on Qwen-3B) and AQuA (34.6 vs.\ 32.3 on Qwen-1.5B), while PaFi performs slightly better on MultiArith (99.0 vs.\ 98.5 on Qwen-7B) and SingleEq (96.2 vs.\ 95.9 on Qwen-3B). Overall, \sparseft demonstrates greater robustness and generalization across tasks, particularly under constrained parameter budgets. Combining \sparseft or PaFi with LoRA enhances task performance by balancing efficiency with accuracy. 
 
Table~\ref{table:generative_tasks2} highlights the performance of \sparseft and PaFi when used directly on the pre-trained MobileLLaMA-2.7B model. The fully fine-tuned MobileLLaMA model achieves 56.8\% accuracy on average. With a 2.7M budget (0.1\% of the entire model), \sparseft recovers 64\% of the performance (achieving 36.3\% accuracy), whereas PaFi recovers 62\% of the average performance. Surprisingly, on more challenging tasks like AQuA and SVAMP, the recovery is higher with both the methods, 87\% with PaFi and 83\% with \sparseft. At a lower budget, the recovery drops for both methods, with \sparseft remaining more robust (recovery 53\%) than PaFi (recovery 42\%). These results indicate that even for larger models (over 1B parameters), full fine-tuning can be avoided with selective alternatives, incurring only slight drops in performance.

\begin{figure*}
\centering
    \includegraphics[width=\textwidth]{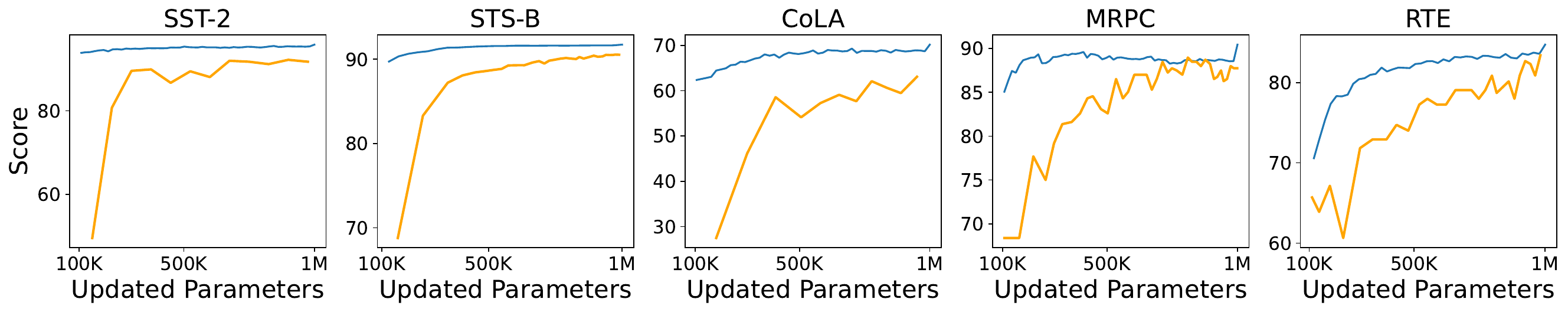}
    \caption{Performance of \heuristic with increment-S (blue line) and repeat-S (orange line) parameter selection. }
    \label{fig:paramwise_score}
\end{figure*}

\section{Analysis}
Here, we study different aspects of \sparseft\ and their importance in the efficient fine-tuning of LLMs. 
\subsection{Importance of Incremental Selection}
We explore a variant of \sparseft\ that uses the repeat-S strategy instead of increment-S. 
As shown in Figure~\ref{fig:paramwise_score}, the increment-S strategy works better for almost all budgets between 100K (sparsity 99.9\%) and 1M (sparsity 98.8\%). Although the performance gap between increment-S and repeat-S reduces for higher budgets, the practical application of the repeat-S strategy remains restricted due to its inferior performance at lower budgets. For a fixed budget, repeat-S typically updates more unique parameters in the model (due to the aggressive exploration strategy at each step) than increment-S. Therefore, it is prone to updating unimportant parameters, leading to lesser performance. Further, for tasks like MRPC and RTE, with limited training samples, repeat-S performance fluctuates across consecutive steps. \sparseft\ on the other hand, minimizes unnecessary parameter updates, achieving a better overall performance.

\begin{figure}
\centering
    \hspace{-22pt}
    \includegraphics[scale=0.48]{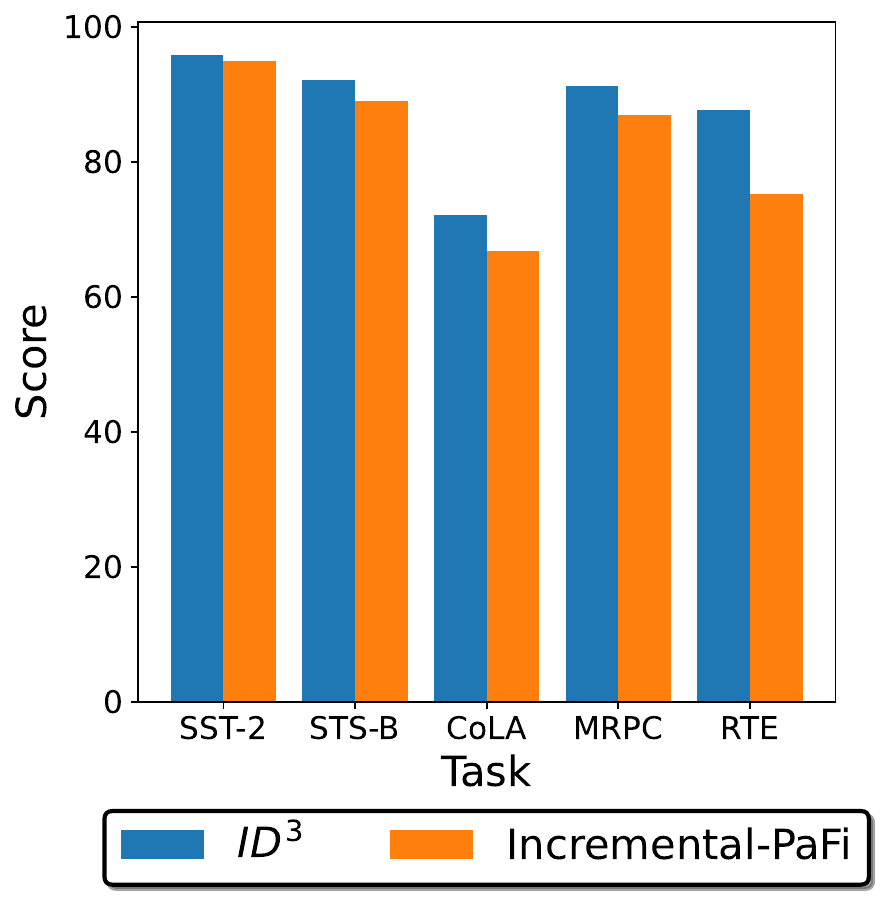}
    \caption{Performance of \sparseft and PaFi with increment-S strategies with DeBERTa-v3.}
    \label{fig:heuristic_analysis}
\end{figure}

\subsection{Importance of the \heuristic Metric}
\label{analysis:importance_d3}

\begin{figure*}[!h]
    \centering
    \subfloat[Performance of \sparseft with different $\epsilon$]{\includegraphics[width=\linewidth]{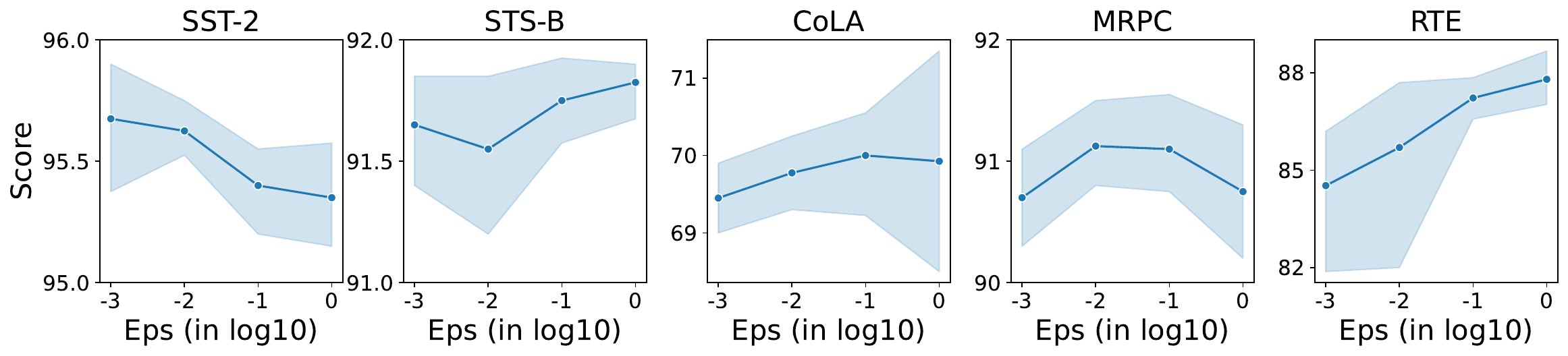}\label{fig:epswise_scores}}
    \quad
    \subfloat[Performance of \sparseft with different $exp$]
    {\includegraphics[width=\linewidth]{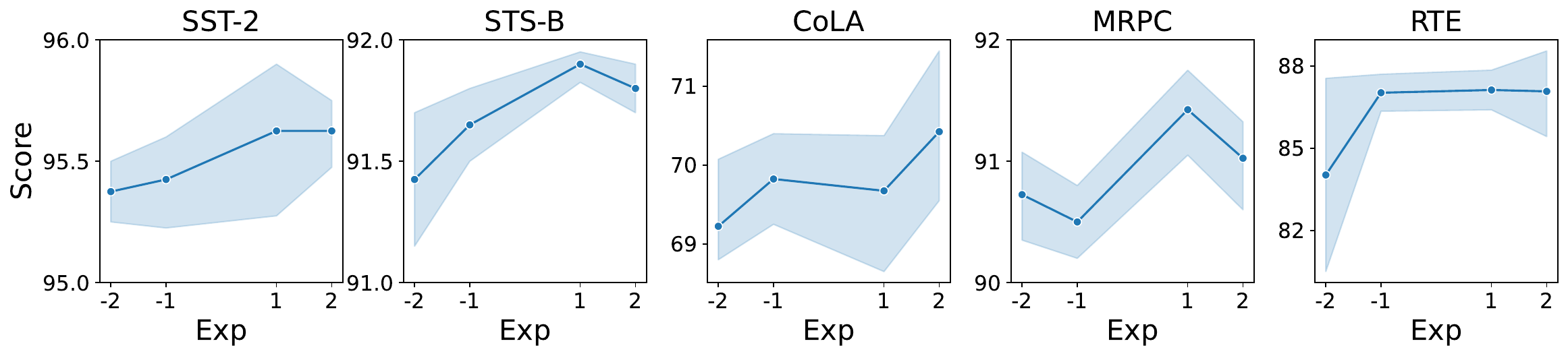}\label{fig:expwise_scores}}
    \caption{Performance of \sparseft\ under different $\epsilon$ and $exp$ values with DeBERTa-v3 backbone model.}
\end{figure*}

Figure~\ref{fig:heuristic_analysis} illustrates the performance of the PaFi heuristic with increment-S selection. On average, using the PaFi heuristic results in a 5\% drop in performance compared to the \heuristic metric, with the largest drop being 12\% on the RTE task. This underwhelming performance highlights the critical role of the \heuristic metric in determining parameter importance during fine-tuning. Unlike \heuristic, the PaFi metric relies solely on the magnitude of parameters to assess importance, potentially overlooking their relative significance towards the task-specific learning objective. This limitation becomes more pronounced when paired with an incremental scheduling strategy. In contrast, \heuristic incorporates both magnitude and gradient information, capturing both the absolute and relative importance of parameters, thereby leading to superior performance.

\begin{table}[!htb]
\captionsetup{font=normalsize}
\centering
\resizebox{1\columnwidth}{!}{
\begin{tabular}{lccccc}
\toprule
& SST-2 & STS-B & CoLA  & MRPC  & RTE \\
\midrule
$\epsilon$   &  1.07 (0.40)  &  0.96  (0.46) &  0.36 (0.83) &  1.38 (0.28) &  1.26 (0.33) \\
$exp$   &  1.15 (0.37) & \textbf{3.02 (0.05)} &  0.51 (0.73) &  0.82 (0.53) &  0.98 (0.44) \\
\bottomrule
\end{tabular}}
\caption{One-way ANOVA test results for assessing the importance of $\epsilon$ and $exp$ values. We report the F-statistics and the p-values. Statistically significant results (p-value  $\leq$ 0.05) are shown in \textbf{bold}.}
\label{tab:exp_eps_anove}
\end{table}

To further understand how different components of \heuristic work, we perform an ablation study on $\epsilon$ and $exp$. Figure~\ref{fig:epswise_scores} highlights that the best performance is achieved typically with $\epsilon \in \{0.1, 1\}$. Lower values of $\epsilon$ have a less smoothing effect, preventing parameters with low gradients from being unmasked unfairly. An interesting trend is also observed with $exp$ (c.f.\ Figure~\ref{fig:expwise_scores}), where $exp \in \{1, 2\}$ consistently performs better than $\{-2, -1\}$. It is, however, worth noting that these performance improvements are statistically insignificant (c.f.\ Table~\ref{tab:exp_eps_anove}). Our one-way ANOVA test highlights that the exact values of $\epsilon$ and $exp$ do not change the overall performance of \sparseft. These results emphasize the robustness of \sparseft under different choices of $\epsilon$ and $exp$ values, demonstrating that \sparseft does not require extensive tuning.

\subsection{Sparsity and importance with \sparseft} 

\begin{figure*}
    \centering
    \subfloat[Tensor Sparsity]{\includegraphics[width=\linewidth]{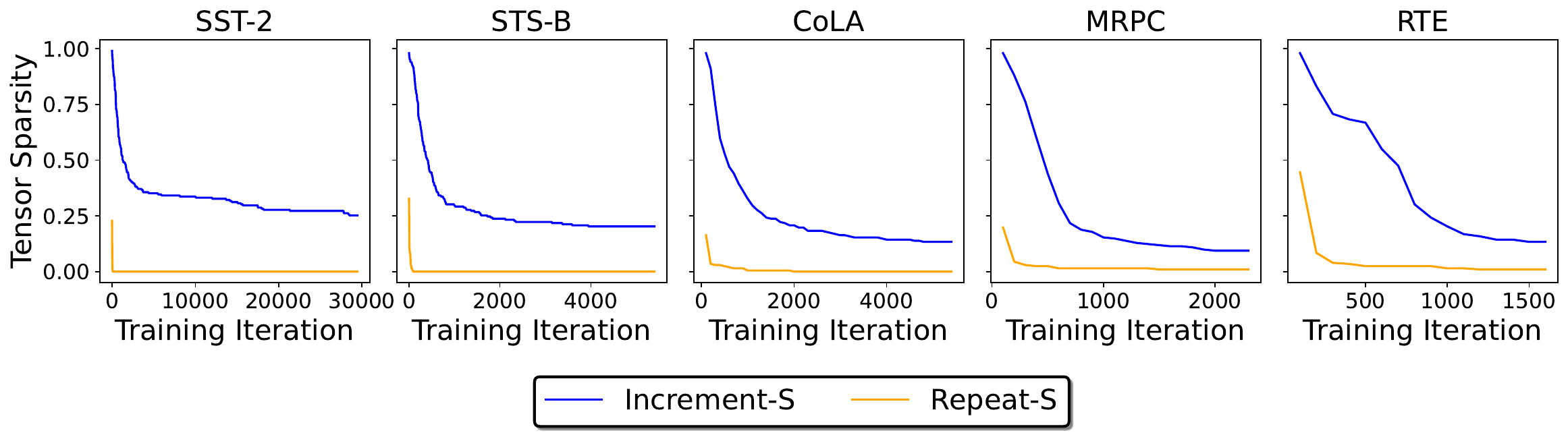}
    \label{fig:sparsity_analysis}}
    \quad
    \subfloat[Tensor Entropy]{\includegraphics[width=\linewidth]{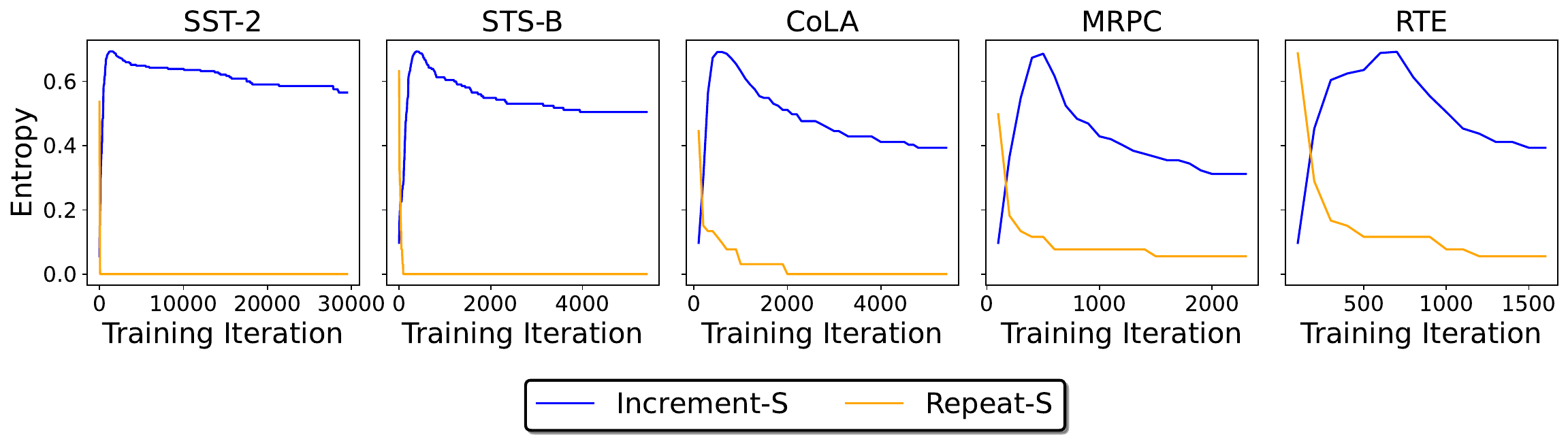}
    \label{fig:entropy_analysis}}
    \caption{Tensor sparsity and entropy with increment-S and repeat-S selection strategies.}
    \label{fig:analysis_sparsity_entropy}
\end{figure*}

For a model with $M$ number of tensor parameters $\{P^i\}_{i=1}^{M}$ fine-tuned with $t$ steps, we define `tensor sparsity' as the number of parameters $P^i$ such that $P^i \cap \Lambda_t = \emptyset$. Figure~\ref{fig:sparsity_analysis} highlights the tensor sparsity for \sparseft with increment-S and repeat-S selection at different training iterations. For all the tasks, tensor sparsity remains close to one for \sparseft\ at the beginning. As the training continues, the tensor sparsity reduces as more scalar parameters are explored. However, the reduction in tensor sparsity stabilizes after a few training steps, indicating more exploitation from the same tensor parameters. A similar behavior is also observed with repeat-S parameter selection. However, with this approach, the tensor sparsity remains much lower, as this selection method exceeds the budget and can potentially fine-tune the entire model.

To understand how \sparseft impacts different tensor parameters in a model, we compute the selection probability for each tensor parameter $P_j$ as $|P_j \cap \Lambda_T|/|P_j|$. Using this probability distribution over all the tensor parameters, we calculate the selection entropy of the fine-tuned model. A high entropy indicates uniform selection probability across different parameters, indicating uniform parameter importance. Figure~\ref{fig:entropy_analysis} suggests that for increment-S, initially, the entropy increases, indicating more exploration of important scalar parameters from different tensor parameters. However, after a few training iterations, the model performs more exploitation by selecting scalar parameters from the same tensor parameters. On the other hand, a repeat-S strategy performs drastic exploration, unmasking most of the tensor parameters quickly and thereby reducing entropy rapidly. 

 \begin{figure*}[t]
     \captionsetup{font=normalsize}
     \centering
     \subfloat[Correlation between FFT and \sparseft parameters]{\includegraphics[width=0.45\linewidth]{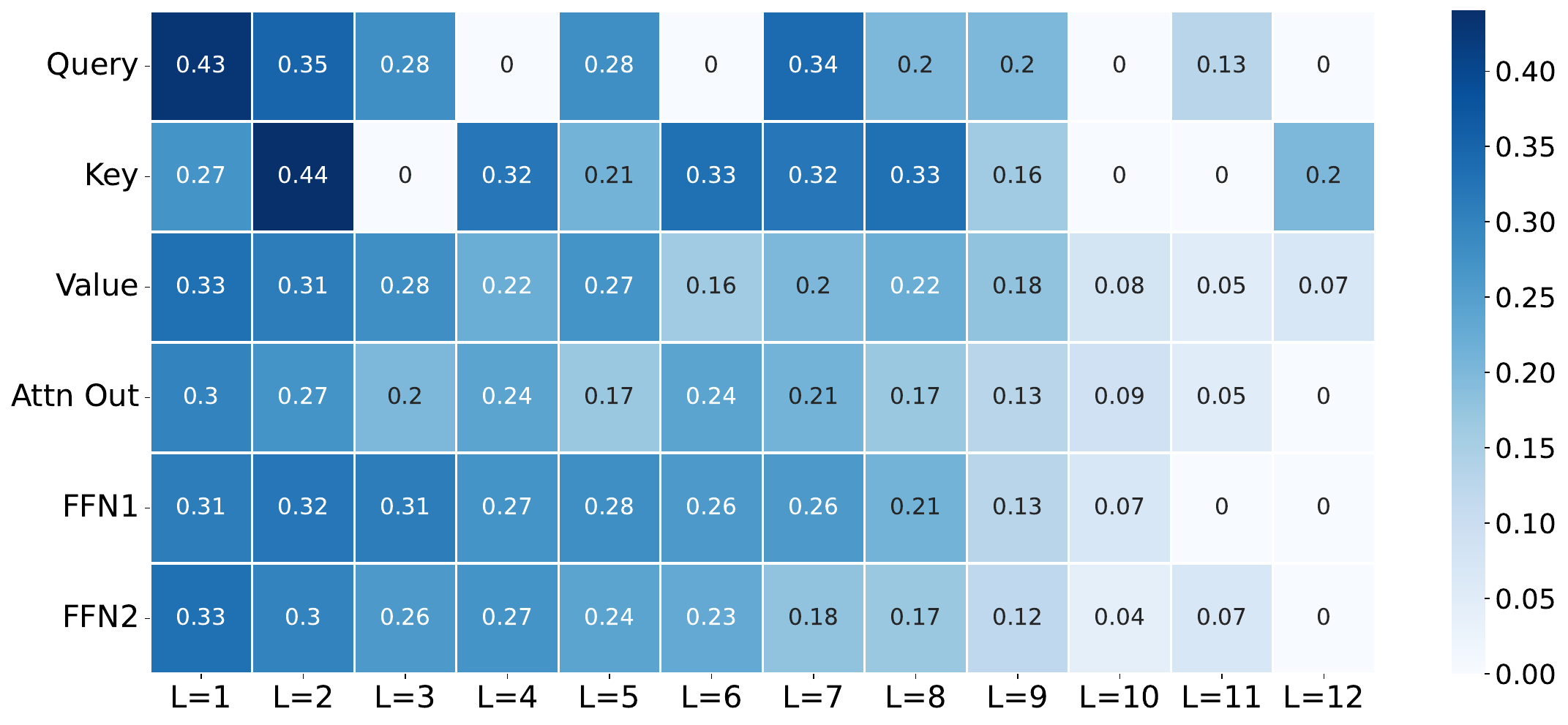}
     \label{fig:stsb_analysis1}}
     \quad
     \subfloat[Importance of parameters unmasked in \sparseft]{\includegraphics[width=0.45\linewidth]{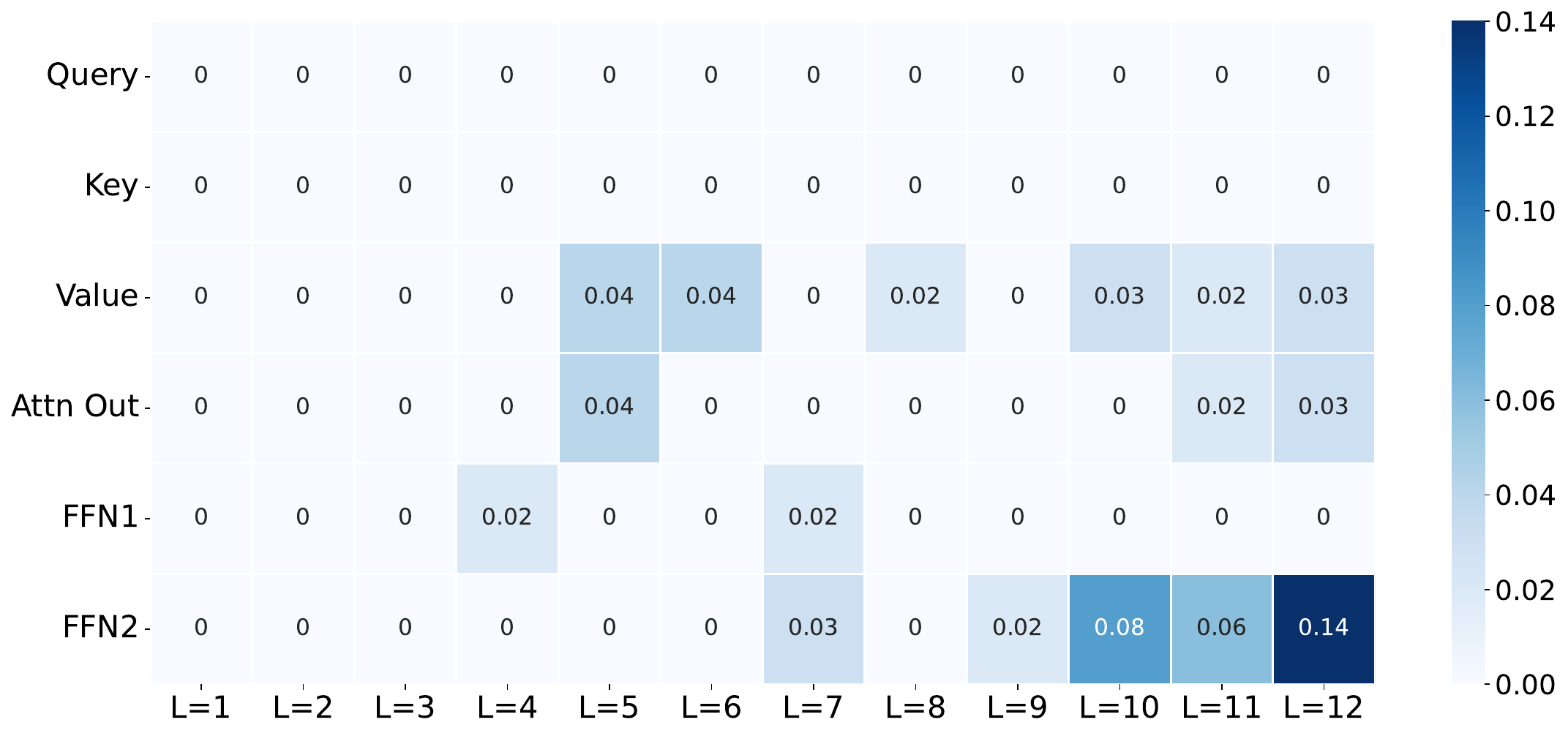}
     \label{fig:stsb_analysis2}}
     \quad
     \subfloat[Parameter update magnitude in FFT and \sparseft]{\includegraphics[width=\linewidth]{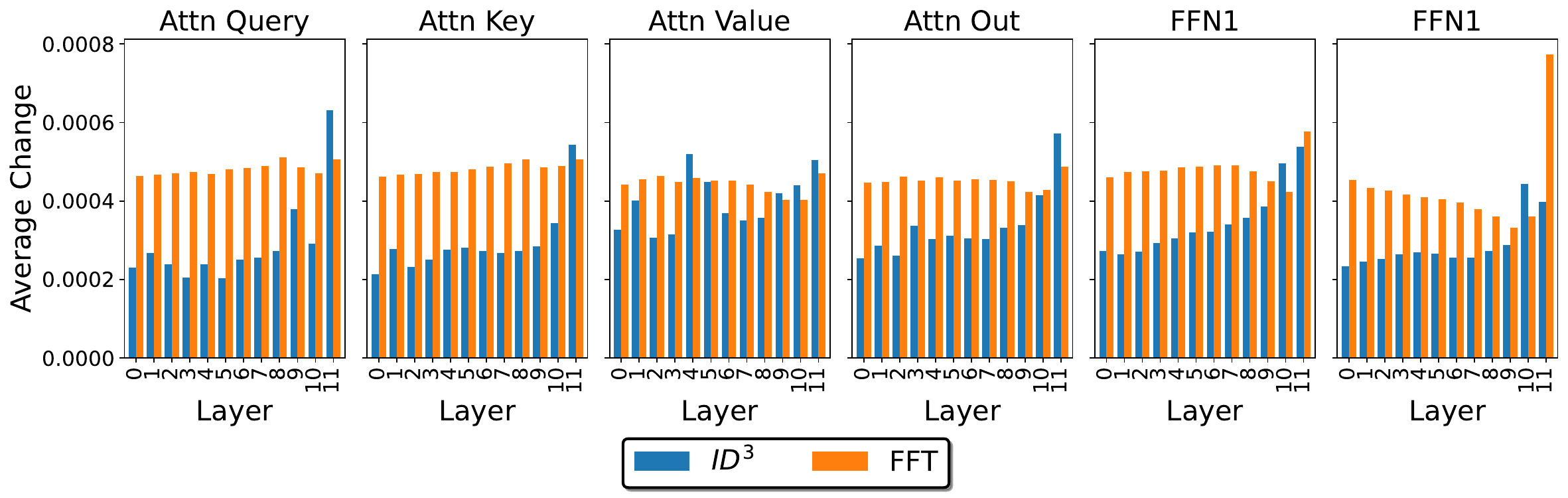}
     \label{fig:stsb_analysis3}}
     \caption{FFT and \sparseft analysis on STS-B. (a) Delta change in parameter weight remains highly correlated between FFT and \sparseft. (b) However, the parameters not selected in \sparseft (potentially unimportant) are also significantly updated with the FFT strategy. (c) At the tensor level, FFT updates the parameters with a higher magnitude than \sparseft. However, unlike FFT, \sparseft incorporates tensor importance and updates the tensors parameter accordingly. \vspace{10pt}}
     \label{fig:stsb_analysis}
 \end{figure*}

\subsection{Difference between FFT and \sparseft}

We perform detailed analysis on the DeBERTa-v3 model on the STS-B task fine-tuned with FFT and \sparseft. The primary objective of this analysis is to gather more insight into the workings of \sparseft and how it behaves compared to full fine-tuning. 

Figure~\ref{fig:stsb_analysis1} shows the Spearman correlation coefficient between the magnitude of parameter change with FFT and \sparseft on the intersecting parameters (\textit{i.e.,}\ parameters updated by both \sparseft and FFT). A high correlation indicates that the parameters common to FFT and \sparseft have similar ordering in terms of importance. On the other hand, Figure~\ref{fig:stsb_analysis2} suggests that under FFT, even the non-overlapping parameters are also subjected to significant gradient updates. This trend highlights the inability of the FFT strategy to determine parameter importance during fine-tuning. Figure~\ref{fig:stsb_analysis3} shows the average change in parameter values after fine-tuning. On average, FFT makes more changes to the parameter values than \sparseft, potentially also updating the unimportant parameters. However, it is worth noting that the magnitude of parameter updates under \sparseft varies between different modules and layers. Self-attention query and key matrices, often considered important for syntactic language understanding, are updated moderately with \sparseft compared to FFT. On the other hand, self-attention value and feed-forward modules that are responsible for capturing semantics and task-specific knowledge are subjected to higher updates with \sparseft. Another interesting observation is that \sparseft makes more change to the later layers of the encoder backbone model, indicating the assignment of greater weight (and hence importance) toward these layers. These demonstrations support the literature~\cite{jawahar-etal-2019-bert, clark-etal-2019-bert} that have previously shown that semantic understanding of language models tends to benefit most from the middle layers, wherein the upper layers contribute more to task-specific feature learning. 

\subsection{Efficiency comparison}
\begin{table*}[!ht]
    \captionsetup{font=normalsize}
    \centering
    \resizebox{1.5\columnwidth}{!}{
    \begin{tabular}{l|c|ccc}
        \toprule
        Method & Peak memory (GB) & Initialization (s) & Update (s) & Overall (s) \\ 
        \midrule
        FFT & 10.10 & 0.0 & 0.00 & 0.23 \\
        BitFit & 10.29 & 0.05 & 0.00 & 0.24 \\
        PaFi & 10.29 & 4.58 & 0.00 & 0.24 \\
        \sparseft & 12.92 & 2.30 & 0.10 & 0.33 \\
        \bottomrule
    \end{tabular}}
    \caption{Computational complexity of selective methods with DeBERTa-v3 model. We report the peak GPU memory consumed (in GB), along with the time taken  in seconds for mask initialization, mask update and overall time during one optimization step.} 
    \label{tab:computation_2}
\end{table*}
Table~\ref{tab:computation_2} compares GPU memory usage and execution time per step for FFT and selective methods. Selective methods like BitFit and PaFi have minimal overhead due to static-S strategies, with memory usage only slightly higher than FFT (10.29 vs.\ 10.10 GB). In contrast, \sparseftnospace’s incremental strategy, requiring additional tensor operations during mask updates, increases memory usage to 12.92 GB. BitFit’s simple parameter selection is the fastest (0.05s), while PaFi’s more complex logic takes twice the time, and \sparseft, with continuous updates, has the slowest per-step execution. However, \sparseft has fixed time and memory costs per step, allowing these overheads to be amortized with larger batch sizes, thereby minimizing their relative impact on the computational cost of fine-tuning.

\vspace{-5pt}
\section{Conclusion}

In this paper, we introduced \sparseft, a novel PEFT technique using incremental-masking-based parameter selection to enhance the fine-tuning of large language models. \sparseft dynamically evaluates and updates parameter importance, effectively balancing exploration and exploitation. Our extensive evaluations showed that \sparseft significantly outperforms traditional PEFT methods, with significantly less number of gradient updates. Additionally, \sparseft integrates seamlessly with other PEFT methodologies, showcasing its versatility. We provide an open-source toolkit with four selective PEFT techniques to support reproducibility and further research. This study marks a significant advancement in PEFT, improving performance and enabling broader scalability of LLMs.

\paragraph{Limitations and future scope} 
While selective PEFT methods do reduce the number of gradient updates (with \sparseft\ achieving competitive performance in half as many updates), the current implementation does not fully leverage this efficiency due to limitations in low-level C++ libraries, which predominantly support dense updates. In order to overcome this, future work will aim to integrate our method directly into the PyTorch library at a lower level, which could better realize the theoretical speedup discussed. We also hope our work inspires further research into the mechanistic activation of selectively updated parameters to deepen the understanding of selective fine-tuning and improve explainability in LLMs.

\clearpage

\bibliography{tacl2021}
\bibliographystyle{acl_natbib}

\clearpage

\section{Appendix}
\subsection{Datasets} \label{app:datasets}

\subsubsection*{Natural language understanding}
For NLU, we evaluate all methods using the following eight tasks from the GLUE benchmark: 

\begin{itemize}
\item RTE (Recognizing Textual Entailment) ~\cite{giampiccolo-etal-2007-third}: Each example consists of two sentences, and the task is to predict whether the second sentence entails the first.

\item MRPC (Microsoft Research Paraphrase Corpus) ~\cite{dolan-brockett-2005-automatically}: The goal is to determine semantic equivalency between the two input sentences.

\item CoLA (Corpus of Linguistic Acceptability) ~\cite{warstadt2019neural}: The task is to predict whether the given sentence is linguistically acceptable.

\item STS-B (Semantic Textual Similarity Benchmark) ~\cite{cer-etal-2017-semeval}: The task is to predict similarity of the given two sentences on a scale of 1 to 5.

\item SST-2 (Stanford Sentiment Treebank) ~\cite{socher-etal-2013-recursive}: The task is to predict whether the sentiment of a given movie review is positive or negative.

\item QNLI (Question-answering NLI) ~\cite{rajpurkar-etal-2016-squad}: Each example consists of a question and a context. The task is to predict whether the given context contains the answer to the question.

\item QQP (Quora Question Pairs) ~\cite{wang2018glue}: The task is to determine whether the questions in the given pair are semantically equivalent.

\item MNLI (Multi-Genre Natural Language Inference) ~\cite{williams-etal-2018-broad}: The task is to determine the relationship between a given premise and hypothesis by predicting whether the premise entails the hypothesis, contradicts it, or neither. This dataset has two validation sets: matched (in-domain) and mismatched (cross-domain) data.
\end{itemize}

\subsubsection*{Token classification}

For token classification, we use the shared task of CoNLL2003 ~\cite{tjong-kim-sang-de-meulder-2003-introduction} that focuses on language-independent named-entity recognition. The goal is to classify each token into four entities: persons, locations, organizations and miscellaneous entities that do not belong to the previous three groups.

\subsubsection*{Summarization}

For summarization, we use the CNN/Daily Mail dataset ~\cite{see-etal-2017-get}, which consists of 300K unique news articles and their highlights written by journalists at CNN and the Daily Mail.

\subsubsection*{Generative reasoning}

For math reasoning tasks, we fine-tune the model using the Math10K dataset and evaluate the final model on the test-split of the following six datasets:

\begin{itemize}
\item GSM8K ~\cite{cobbe2021training}: This dataset contains diverse grade school math word problems. The task is to perform a sequence of elementary calculations to obtain the final answer.

\item SVAMP ~\cite{patel2021nlp}: This dataset is created by introducing straightforward variations to single-unknown arithmetic word problems designed for grade levels up to 4.

\item MultiArith ~\cite{roy2015solving}: This dataset consists of multi-step arithmetic word problems involving basic operations, such as addition followed by subtraction or subtraction followed by division.

\item AddSub ~\cite{hosseini2014learning}: This corpus contains arithmetic problems with addition and subtraction.

\item AQuA ~\cite{ling2017program}: This dataset contains algebraic word problems along with answer rationales.

\item SingleEq ~\cite{koncel2015parsing}: This dataset contains sentences expressing mathematical relations that form a single equation.

\item Math10K ~\cite{hu2023llm}: This dataset was constructed by combining training examples from GSM8K, AQuA, MAWPS and MAWPS-single ~\cite{koncel-kedziorski-etal-2016-mawps}. The original datasets contained only equations and final answers. To enhance them with explanations, the authors employed ChatGPT to generate reasoning steps for each example, creating the final Math10K dataset.

\end{itemize}

The train, validation and test splits of all the datasets are shown in Table~\ref{tab:datasets_splits}.

\begin{table*}
\centering
\resizebox{1.16\columnwidth}{!}{%
\begin{tabular}{llccc}
\toprule
Domain                & Dataset     & \# train & \# validation & \# test \\ \midrule
\multirow{9}{*}{GLUE}   
   & RTE         & 2.5K     & 277           & 3K      \\ 
   & MRPC        & 3.7K     & 408           & 1.7K       \\ 
   & CoLA        & 8.5K     & 1K            & 1K       \\ 
   & STS-b       & 5.7K     & 1.5K          & 1.4K       \\ 
   & SST-2       & 67K      & 872           & 1.8K       \\ 
   & QNLI        & 105K     & 5.5K          & 5.5K       \\ 
   & QQP         & 364K     & 40K           & 390K       \\ 
   & MNLI         & 393K     & 10K           & 10K       \\ \midrule
NER  & CoNLL2003         & 14K      & 3.2K          & 3.5K       \\ \midrule
Summarization & CNN DailyMail & 287113 & 13368 & 11490 \\ \midrule
\multirow{9}{*}{Math Reasoning}
        & Math10k    & 10K      & -             & -       \\ 
        & GSM8K       & 8.8K     & -             & 1319    \\ 
        & SVAMP       & -        & -             & 1000    \\ 
        & MultiArith  & -        & -             & 600     \\ 
        & AddSub      & -        & -             & 395     \\ 
        & AQuA        & 100K     & -             & 254     \\ 
        & SingleEq    & -        & -             & 508     \\ 
\bottomrule
\end{tabular}}
\caption{Datasets and data splits for different tasks used in the paper.}
\label{tab:datasets_splits}
\end{table*}

\newcommand{\common}{%
\begin{tabular}
{llcccc}
\toprule
& Category & NLU  & NER & Summarization & Math Reasoning \\ \midrule
PEFT Method & hyperparameter & All tasks & CoNLL2003 & CNN/Daily Mail & All tasks \\ \midrule
\multirow{6}{*}{All methods} & batch size & $16$ & $16$ & $64$ & $4$ \\ 
 & \multirow{4}{*}{learning rate} & $1\times10^{-4}$ & $1\times10^{-4}$ & \multirow{4}{*}{$1\times10^{-4}$} & \multirow{4}{*}{$3\times10^{-4}$} \\ 
 &  & $3\times10^{-4}$ & $3\times10^{-4}$ & & \\ 
 &  & $5\times10^{-4}$ & $5\times10^{-4}$ & & \\ 
 &  & $7\times10^{-4}$ & $7\times10^{-4}$ & & \\ 
 & seed & $\{ 6, 7, 8, 9 \}$ & $\{ 6, 7, 8, 9 \}$ & $9$ & $42$ \\ \midrule
\multirow{4}{*}{FFT} & \multirow{4}{*}{learning rate} 
& $5\times10^{-6}$ & $5\times10^{-6}$ & \multirow{4}{*}{$1\times10^{-5}$} & \multirow{4}{*}{-} \\
& & $7\times10^{-6}$ & $7\times10^{-6}$ & & \\
& & $1\times10^{-5}$ & $1\times10^{-5}$ & & \\
& & $3\times10^{-5}$ & $3\times10^{-5}$ & & \\ \midrule
  
\multirow{2}{*}{\sparseft} & $exp$ & $2$ & $2$ & $1$ & $\{0, 1\}$ \\ 
 & $\epsilon$ & $1$ & $1$ & $1\times10^{-3}$ & $1$ \\ \midrule
\multirow{3}{*}{Fish} & $\text{num\_samples}$ & $1024$ & $1024$ & \multirow{3}{*}{-} & \multirow{3}{*}{-} \\ 
 & $\text{sample\_type}$ & "label" & "label" & & \\ 
 & $\text{grad\_type}$ & "square" & "square" & & \\ \midrule
\multirow{8}{*}{LoRA} & $\text{lora\_r}$ & $8$ & \multirow{8}{*}{-} & \multirow{8}{*}{-} & $\{2, 8, 32\}$ \\ 
 & $\text{lora\_alpha}$ & $8$ &  & & $\{16, 64\}$ \\ 
 & \multirow{6}{*}{$\text{lora\_modules}$} & query\_proj & &  & query\_proj \\ 
 &  & key\_proj &  & & key\_proj \\ 
 &  & value\_proj &  & & value\_proj \\
 &  & attention.output.dense & & & up\_proj \\
 &  & intermediate.dense & & & down\_proj \\
 &  & output.dense & & &  \\ \bottomrule
\end{tabular}}

\newcommand{\task}{%
\begin{tabular}{llcccc}
\toprule
Benchmark & Dataset   & Metric                                     & Epochs & Eval\_Steps & Max Seq Length \\ \midrule
\multirow{8}{*}{GLUE}      & RTE       & Accuracy                                   & $30$   & $100$       & $256$          \\ 
      & MRPC      & Accuracy                                   & $30$   & $100$       & $256$          \\ 
      & CoLA      & Matthews Correlation                       & $20$   & $200$       & $256$          \\ 
      & STS-B     & Avg of Spearman and Pearson Corr.    & $15$   & $200$       & $256$          \\ 
      & SST-2     & Accuracy                                   & $7$    & $500$       & $256$          \\ 
      & QNLI      & Accuracy                                   & $7$    & $1000$      & $256$          \\ 
      & QQP       & Accuracy                                   & $3$    & $4000$      & $256$          \\ 
      & MNLI      & Accuracy                                   & $3$    & $4000$      & $256$          \\ \midrule
NER       & CoNLL2003 & F1                                         & $20$   & $300$       & $384$          \\ \midrule

\multirow{2}{*}{Summarization} & \multirow{2}{*}{CNN/Daily Mail} & \multirow{2}{*}{Rouge1/Rouge2/RougeL} & \multirow{2}{*}{3} & \multirow{2}{*}{1000} & Source Length = 512 \\ 
& & & & & Target Length = 128 \\
\midrule

\multirow{6}{*}{Math Reasoning} & Math10K & - & $3$ & - & $256$ \\
 & GSM8K & Accuracy & - & $80$ & $256$ \\
 & SVAMP & Accuracy & - & $80$ & $256$ \\
 & MultiArith & Accuracy & - & $80$ & $256$ \\
 & AddSub & Accuracy & - & $80$ & $256$ \\
 & AQuA & Accuracy & - & $80$ & $256$ \\
 & SingleEq & Accuracy & - & $80$ & $256$ \\ \bottomrule
\end{tabular}
}

\begin{table*}[h!]
  \centering
  \subfloat[Common and PEFT method specific hyperparameters]{\resizebox{1.8\columnwidth}{!}{\common} \label{tab:hyperparameters_method_2}}%
  \qquad
  \subfloat[Task specific hyperparameters]{\resizebox{1.8\columnwidth}{!}{\task} \label{tab:hyperparameters_glue_task_specific}}
  \caption{Details of all the hyperparameters used in the paper.}%
  \vspace{10cm}
  \label{tab:hyperparameters_all}
\end{table*}

\subsection{Hyperparameters} 
\label{app:hyperparameters}
All the common and task-specific hyperparameters are shown in Table~\ref{tab:hyperparameters_method_2} and~\ref{tab:hyperparameters_glue_task_specific}, respectively. 

\subsubsection*{Common hyperparameters}  
\paragraph{Budget}  
For NLU and NER tasks, we use parameter budgets of 103K and 320K. The 103K budget is selected to align with the number of parameters fine-tuned using BitFit, which updates only the model's bias terms. The 320K budget is chosen to reduce the performance gap between PEFT methods and full fine-tuning.  
For summarization tasks, we adopt budgets of 100K, 320K, and 1M. These choices align with the NLU task budgets while also including a larger budget for comparative analysis.  
For the mathematical reasoning task, the parameter budget for each model is set to match the number of parameters associated with applying LoRA at a rank of 2 to that model.

\paragraph{Learning rate} For NLU and NER tasks, we use learning rates of around $3 \times10^{-4}$ for selective fine-tuning. For full fine-tuning however, these learning rates are typically too high, and hence we use learning rates of around $7 \times10^{-6}$. These learning rates were selected without bias toward any specific method and following the common practice of choosing rates within the range of $1\times10^{-3}$ to $1\times10^{-4}$. 
For summarization and reasoning tasks, we fine-tune the models with learning rates of $1 \times 10^{-4}$ and $3 \times 10^{-4}$, respectively. 

\paragraph{Scoring}
For NLU and NER tasks, we conducted a single run for each learning rate, resulting in four runs per method. For each run, the maximum score based on the evaluation metric (accuracy or correlation) was recorded. The final score was calculated as the average of the four scores.

\subsubsection*{Specific hyperparameters}

\paragraph{PEFT related}
For \sparseft we demonstrated in Section~\ref{analysis:importance_d3} that better results were achieved with positive values of $exp$. The parameter $\epsilon$ acts as a smoothing factor, with smaller values generally yielding improved outcomes. As outlined in the Fish mask paper ~\cite{sung2021training}, the optimal hyperparameters for ``num\_samples,'' ``sample\_type,'' and ``grad\_type'' were used. Meanwhile, BitFit and PaFi do not use any hyperparameters.

\newcommand{\stdgluefinal}{%
\begin{tabular}{llcccccccccc}
        \toprule
        Budget & Method & MNLI-m & MNLI-mm & QQP & QNLI & SST-2 & STS-B & CoLA & MRPC & RTE & Avg \\
        \midrule
        184M & Full-FT & 0.30 & 0.40 & 0.36 & 0.29 & 0.36 & 0.50 & 1.67 & 1.00 & 1.83 & 0.74 \\
        \hdashline
        \multirow{5}{*}{103K} 
         & R-Mask & 4.89 & 4.48 & 2.03 & 3.05 & 1.74 & 7.49 & 3.39 & 6.25 & 8.58 & 4.66 \\
         & Fish & 0.48 & 0.51 & 0.29 & 0.43 & 0.33 & 0.79 & 2.17 & 1.10 & 2.90 & 1.00 \\
         & PaFi & 0.82 & 0.62 & 0.73 & 0.89 & 0.62 & 1.80 & 1.80 & 1.80 & 2.51 & 1.29 \\
         & BitFit & 1.08 & 0.75 & 0.70 & 0.89 & 0.51 & 1.67 & 1.09 & 1.42 & 2.91 & 1.23 \\
         & \sparseft & 0.12 & 0.13 & 0.27 & 0.17 & 0.25 & 0.27 & 0.45 & 0.42 & 3.14 & 0.58 \\
        \hdashline
        \multirow{4}{*}{320K} 
         & R-Mask & 2.00 & 1.58 & 1.27 & 1.96 & 1.27 & 5.33 & 2.82 & 8.88 & 5.13 & 3.36 \\
         & Fish & 0.12 & 0.11 & 0.45 & 0.13 & 0.16 & 0.36 & 1.39 & 0.47 & 2.45 & 0.63 \\
         & PaFi & 0.92 & 0.92 & 0.79 & 0.46 & 0.60 & 1.07 & 1.41 & 0.82 & 2.57 & 1.06 \\
         & \sparseft & 0.29 & 0.38 & 0.27 & 0.11 & 0.11 & 0.28 & 1.45 & 0.37 & 1.60 & 0.54 \\
        \bottomrule
    \end{tabular}}

\newcommand{\stdlorafinal}{%
\begin{tabular}{llcccccccccc}
\toprule
Budget & Method             & MNLI-m & MNLI-mm & QQP  & QNLI & SST-2 & STS-B & CoLA & MRPC & RTE & Avg    \\ \hline
1.33M & LoRA (r=8)          & 0.23   & 0.12    & 0.12 & 0.36 & 0.21  & 0.29  & 1.42 & 1.32 & 0.86 & 0.55 \\
\hdashline
320K & PaFi + LoRA (r=8)    & 0.36 & 0.48 & 0.86 & 0.45 & 0.09 & 0.60 & 0.74 & 0.70 & 1.77 & 0.67 \\
320K & \sparseft + LoRA (r=8) & 0.35 & 0.28 & 0.26 & 0.08 & 0.15 & 0.22 & 0.66 & 0.42 & 1.33 & 0.42 \\
\bottomrule
\end{tabular}
}

\newcommand{\stdnerfinal}{%
\begin{tabular}{lcccccc}
        \toprule
        Budget & Full-FT & Fish & PaFi & BitFit & \sparseft & R-Mask \\ \midrule
        103K & - & 0.41 & 1.66 & 1.69 & 0.64 & 31.09 \\ 
        320K & - & 0.16 & 1.25 & - & 0.28 & 7.70 \\ 
        184M & 0.20 & \multicolumn{5}{c}{-} \\ 
        \bottomrule
\end{tabular}
}

\begin{table*}[h!]
  \centering
  \subfloat[Standard deviations corresponding to the results in Table \ref{tab:glue_tasks}.]{\resizebox{2\columnwidth}{!}
  {\stdgluefinal}\label{tab:std_glue}}%
  \qquad
  \subfloat[Standard deviations corresponding to the results in Table \ref{tab:glue_lora}.]{ \resizebox{2\columnwidth}{!}{\stdlorafinal}\label{tab:std_glue_lora}}
  \qquad
  \subfloat[Standard deviations corresponding to the results in Table \ref{tab:ner_2}.]{\resizebox{1.1\columnwidth}{!}{\stdnerfinal}\label{tab:std_ner}}
  \caption{Standard deviations corresponding to GLUE and NER results.}%
  \label{tab:glue_std_final}
\end{table*}

\paragraph{Task related}
The number of epochs, evaluation steps, and maximum sequence length are determined based on the size of the training and evaluation datasets. These details are presented in Table~\ref{tab:hyperparameters_glue_task_specific}.

\subsection{Standard deviations} 
We report the standard deviations on GLUE and NER tasks for all the baselines in Table~\ref{tab:glue_std_final}.

\label{app:std_deviations}

\subsection{Number of runs}

We report the number of runs (and the different hyperparameters used in these runs) conducted for each experiment in Table~\ref{tab:result_types}.

\begin{table*}[!ht]
    \centering
    \resizebox{10cm}{!}{%
    \begin{tabular}{lcc}
        \toprule
        Table/Figure & Number of runs for each method & Hyperparameter varied \\
        \midrule
        Figure~\ref{fig:glue_results} & 4 & Learning rate\\
        Table~\ref{tab:glue_tasks} & 4 & Learning rate\\
        Table~\ref{tab:glue_lora} & 4 & Learning rate\\
        Table~\ref{tab:reused_mask_results} & 1 & -\\
        Table~\ref{tab:ner_2} & 4 & Learning rate\\
        Table~\ref{tab:summ_mean} & 1 & -\\
        Table~\ref{table:generative_tasks} & 1 & -\\
        Table~\ref{table:generative_tasks2} & 1 & -\\
        Figure~\ref{fig:paramwise_score} & 1 & -\\
        Figure~\ref{fig:heuristic_analysis} & 1 & -\\
        Figure~\ref{fig:epswise_scores} & 4 & Exponent\\
        Figure~\ref{fig:expwise_scores} & 4 & Epsilon\\
        Figure~\ref{fig:analysis_sparsity_entropy} & 1 & -\\
        Figure~\ref{fig:stsb_analysis} & 1 & -\\
        \bottomrule
    \end{tabular}}
    \caption{Number of runs and tuned hyperparameters for different experiments.} 
    \label{tab:result_types}
\end{table*}

\begin{table*}[t!]
    \centering
    \resizebox{14cm}{!}{
    \begin{tabular}{lccccc}
        \toprule
        \bf Table & \bf Tested across & \bf Bootstrapped & \bf Paired method ($Y$) & \bf P-value & \bf Significant\\
        \midrule
        Table~\ref{tab:glue_tasks} & Budget & Yes & Task-wise best baseline & $8\times10^{-7}$ & Yes\\
        Table~\ref{tab:glue_lora} & Task & Yes & PaFi + LoRA & $0.04$ & Yes\\
        Table~\ref{tab:reused_mask_results} & Task & - & SparseAdapter & $0.12$ & No \\
        Table~\ref{tab:ner_2} & Budget & Yes & Fish Mask & $0.01$ & Yes\\
        Table~\ref{tab:summ_mean} & Budget & - & PaFi & $0.13$ & No\\
        Table~\ref{table:generative_tasks} & Model & - & Task-wise best baseline &  $0.06$ & No\\
        Table~\ref{table:generative_tasks2} & Task, budget & - & PaFi & $0.01$ & Yes\\
        \bottomrule
    \end{tabular}
    }
    \caption{Description of all statistical significance tests. For all the tests we use the \textbf{null hypothesis:} \textit{The observation $X_i$ (\sparseft) - $Y_i$ (mentioned in column 3) is symmetric about $\mu = 0$}. We use an \textbf{alternative hypothesis:} \textit{The observations $X_i$ - $Y_i$ are symmetric about $\mu > 0$}. The \textbf{significance level} is set as 0.05. For Tables~\ref{tab:glue_tasks}, \ref{tab:glue_lora} and \ref{tab:ner_2} (where results with multiple hyperparameter configurations are available), we use ``bootstrapping,'' where we compare all the pairwise results obtained by \sparseft and the best baseline for a given configurations.} 
    \label{tab:statistical_test_details}
\end{table*}

\subsection{Statistical significance testing}
\label{appx:stat_test}

We perform statistical tests for all the results reported in the paper. In particular, we perform the Wilcoxon signed-rank test\footnote{Note that in order to use this test, the distribution must be symmetric about a center. We assume that this holds for all tasks.}, a non-parametric test, since the nature of the accuracy distribution with different seeds and learning rates is not known. We perform the paired variant of these tests with data-points in a pair being drawn from the same configuration (task, budget). Table \ref{tab:statistical_test_details} provides for each result the configurations across which the test has been done. Further, in all cases where multiple runs for a given configuration are available (Tables~\ref{tab:glue_tasks}, \ref{tab:glue_lora} and \ref{tab:ner_2}) we use what we call ``bootstrapping,'' where we pair each run of \sparseft with every run of the compared baseline, obtaining $n^2$ pairs (where $n$ is the number of runs for a given configuration for each method). For instance, in the GLUE table (Table~\ref{tab:glue_tasks}) we have two budgets (103K and 320K) and four runs per entry (obtained by varying the learning rate). So, in total we have $4\times4$ pairs for a given budget and in total $2\times4\times4 = 32$ pairs for a task (across the two budgets). As per common practice, we take the significance level as 0.05, which means that a result is considered statistically significant if its p-value < 0.05. For each result table, we take take the best task-wise baseline (excluding full fine-tuning or it's alternatives) as the paired method.

\end{document}